\pdfoutput=1

\documentclass[11pt]{article}

\usepackage{acl}

\usepackage{times}
\usepackage{latexsym}
\usepackage{pifont}

\usepackage[T1]{fontenc}

\usepackage[utf8]{inputenc}

\usepackage{microtype}

\usepackage{inconsolata}

\usepackage{graphicx}

\usepackage{float}
\usepackage{booktabs}
\usepackage{enumitem}
\usepackage{amsmath}
\usepackage{amsfonts}
\usepackage{multirow}
\usepackage{xcolor}
\usepackage{algorithm}
\usepackage{algpseudocode}
\def\Snospace~{Section }

\title{Auto-Prompt Generation is Not Robust: \\
          Prompt Optimization Driven by Pseudo Gradient}

\author{
 \textbf{Zeru Shi\textsuperscript{1}},
 \textbf{Zhenting Wang\textsuperscript{1}},
 \textbf{Yongye Su\textsuperscript{2}},
 \textbf{Weidi Luo\textsuperscript{3}}, \\
 \textbf{Hang Gao\textsuperscript{1}},
 \textbf{Fan Yang\textsuperscript{4}},
  \textbf{Ruixiang Tang\textsuperscript{1}},
 \textbf{Yongfeng Zhang\textsuperscript{1}},
\\
\\
 \textsuperscript{1} Rutgers University,
 \textsuperscript{2} Purdue University,
\\
 \textsuperscript{3} Georgia University,
 \textsuperscript{4} Wake Forest University
\\
}

\begin{document}
\maketitle
\begin{abstract}
While automatic prompt generation methods have recently received significant attention, their robustness remains poorly understood. In this paper, we introduce \textbf{PertBench}, a comprehensive benchmark dataset that includes a wide range of input perturbations, designed to systematically evaluate the robustness of current auto-prompting techniques. Our analysis reveals substantial vulnerabilities in existing prompt generation strategies, where even minor modifications to the prompt can lead to significant differences in model output. To address this issue, we propose \textbf{PGO}, a gradient-free prompt generation framework that leverages perturbation types as pseudo-gradient signals to guide LLMs in producing more robust prompts. In contrast to existing methods that assess prompt quality only on clean, well-structured inputs, our approach explicitly emphasizes robustness under noisy and perturbed conditions. Extensive experiments across diverse tasks and multiple LLMs show PGO consistently outperforms previous methods in maintaining performance under input perturbations.

\end{abstract}

\section{Introduction}

\begin{figure}[ht]
    \centering
    \includegraphics[width=0.45\textwidth]{fig/example.pdf}
    \caption{An example of adding a perturbation to the input and its result in classification task, where the top one indicates that the prompt gets the correct output under normal input, the middle one indicates that the prompt gets the wrong result under perturbed input, the bottom indicates the PGO gets correct result under perturbed input.}
    \label{fig0:head}
\end{figure}
Large Language Models (LLMs) have demonstrated impressive capabilities across a wide spectrum of tasks.~\cite{abburi2023generative, ge2024openagi, laban2023summedits}. Given their broad applicability, a growing body of research has focused on strategies to more effectively elicit and amplify these capabilities. Among them, prompt engineering has emerged as a central technique, with recent advances increasingly relying on LLMs themselves to automate the prompt optimization process. Representative approaches include instruction fine-tuning~\cite{ouyang2022training, ziegler2019fine}, Chain-of-Thought (CoT) prompting~\cite{shum2023automatic, zhang2022automatic}, and inference-guided prompt generation~\cite{wang2023promptagent, liu2024large}. Despite their success on clean, well-structured inputs, these methods share a key limitation: they assume idealized input conditions, and thus fail to account for the noisy, imperfect inputs commonly encountered in real-world scenarios, such as typographical errors, ambiguous phrasing, or factual imprecision. These input perturbations can significantly degrade model performance, revealing a critical robustness gap in current prompt optimization techniques. As shown in Figure.~\ref{fig0:head}, even subtle input corruptions (e.g., a minor typo) can mislead an LLMs and trigger incorrect predictions in tasks like classification.

\begin{figure*}[h]
    \centering
    \includegraphics[width=1\textwidth]{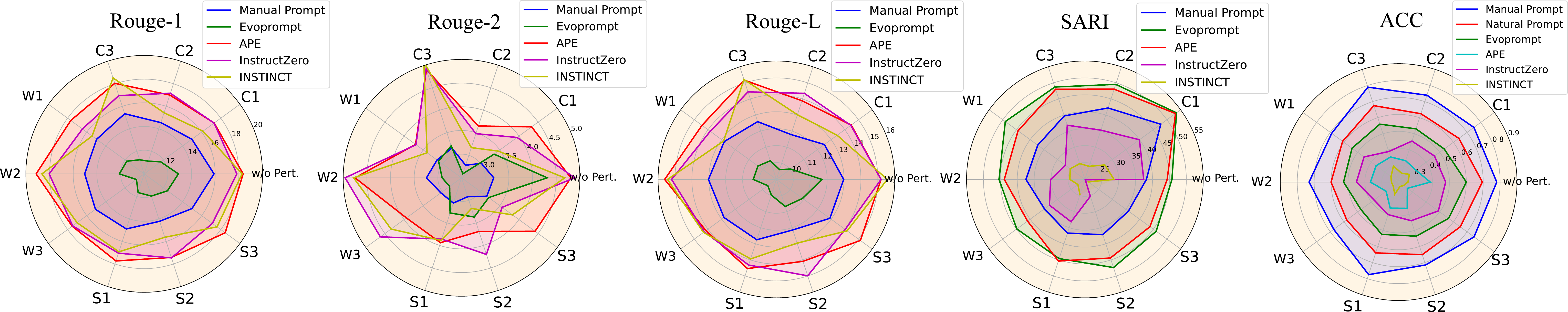}
    \caption{Comparison of the execution results of all different methods on PertBench for the three tasks.}
    \label{fig1:radar}
\end{figure*}

\begin{table}
\centering
\resizebox{0.45\textwidth}{!}{
\begin{tabular}{c|c|c|c}
\hline
\begin{tabular}[c]{@{}c@{}}Perturbation \\ level \end{tabular} & Name. & \multicolumn{1}{c|}{Explanation} & Type \\ \hline
\multirow{3}{*}{Character} & C1 & \begin{tabular}[c]{@{}l@{}}Change words to have typos \end{tabular} & P1\\ 
\cline{2-4} 
 & C2 & Change Letters. & P1\\ 
 \cline{2-4} 
 & C3 & Add extraneous characters & P1\\ 
 \hline
\multirow{3}{*}{Word} & W1 & Change word to synonyms. & P2\\ 
\cline{2-4} 
 & W2 & \begin{tabular}[c]{@{}l@{}} Delete meaningless words.\end{tabular} & P2\\ \cline{2-4} 
 & W3 & Add neutral words. & P2\\ \hline
\multirow{3}{*}{Sentence} & S1 & \begin{tabular}[c]{@{}l@{}}Add meaningless handle\end{tabular} & P1 \\ \cline{2-4} 
 & S2 & Paraphrase the sentence. & P2\\ \cline{2-4} 
 & S3 & Change the syntactic structure. & P2\\ \hline
\end{tabular}}
\caption{Explanation of different kinds of perturbation.}
\label{tab:pt}
\end{table}
In this work, we introduce \textbf{PertBench}, a comprehensive benchmark designed to evaluate the robustness of prompt-based methods under diverse input perturbations. Our benchmark spans three representative natural language processing tasks, text simplification, summarization, and classification—capturing a wide spectrum of practical use cases. In contrast to prior datasets such as NoiseBench~\cite{merdjanovska-etal-2024-noisebench, moradi2021evaluating, dong2023revisit}, which primarily focus on character-level noise within a single task or short form content (e.g., dialogues), PertBench provides broader coverage across multiple granularity levels, including character-level, word-level, and sentence-level perturbations. Moreover, we extend the perturbation setting to long form text and ensure a uniform and comprehensive perturbation strategy across all tasks. The specific categories of perturbations are summarized in Table~\ref{tab:pt}. This design enables a more systematic and rigorous evaluation of prompt robustness across tasks and text lengths. Our results demonstrate that existing prompt generation methods experience substantial performance degradation under perturbed inputs, exposing their vulnerability and limited generalization. 

To mitigate this issue, we propose \textbf{PGO} (Pseudo Gradient Optimization), a gradient-free framework for robust discrete prompt optimization. Unlike conventional approaches that rely on model gradients to update prompts, PGO leverages perturbation information as guidance to guide prompt refinement. We categorize perturbations into two types P1 and P2 depending on whether they cause significant semantic shifts. Correspondingly, PGO introduces two complementary optimization strategies tailored to these categories and iteratively refines prompts via feedback-driven updates. Through this design, PGO effectively enhances prompt robustness against a broad range of input disturbances. Our main contributions are summarized as follows:

\begin{itemize}[left=0pt]
\item We present \textbf{PertBench}, a comprehensive benchmark for evaluating prompt robustness under diverse input perturbations. It includes nine perturbation types across three core task categories for systematic robustness assessment.

\item We propose \textbf{PGO}, a pseudo gradient–based prompt optimization framework that enhances robustness across perturbations without requiring model internals or true gradients.

\item Extensive experiments show that existing methods degrade notably on PertBench, while PGO consistently improves robustness across tasks and perturbation settings.

\end{itemize}

\section{Benchmark Construction}
\subsection{Details of Benchmark}
\subsubsection{Basic Datasets}
In the \textbf{classification tasks}, we focus on six datasets to apply and test perturbations. These include sentiment classification datasets: SST-2~\cite{socher2013recursive}, CR~\cite{hu2004mining}, SST-5~\cite{socher2013recursive}, and MR~\cite{pangb2005exploitingclassrelationshipsforsentimentcate}, as well as topic classification datasets: AG's News~\cite{zhang2015character} and TREC~\cite{voorhees2000building}. In language generation tasks, we utilize the Asset~\cite{alva2020asset} for \textbf{text simplification}, which includes multiple benchmarks for reference translations. For the \textbf{text summarization task}, we use the XSum~\cite{narayan2018don}, which consists of concise summaries generated from longer texts. More details of PertBench in the~\autoref{bench_detail}.

\subsubsection{Perturbation Types}
We adopt a widely used approach following~\citet{xu2023llm}, which form the basis for the perturbations in our experiments (See Table.~\ref{tab:pt}). These disturbances are categorized into two groups: \textit{P1} and \textit{P2}. P1 introduce typographical errors and non-sensical strings into the text, without significantly altering the underlying semantic structure of the sentence. P2 alter the semantic structure of sentences by modifying their syntactic composition, while maintaining their core meaning. These modifications can result in a bias in how LLMs interpret the input, potentially affecting their understanding. We compare the similarity between the perturbed data set and the original data set, and the experiment shows that we complete the perturbation of the data while maintaining the original similarity. The results in the~\autoref{bench_detail}.

\subsection{Benchmark Construction}
When constructing the dataset, we apply nine different perturbations to each type of task to construct separate perturbed datasets. To be more specific, we introduce various types of perturbation-specific guide words as a fixed gradient \( g \), and use it to guide the perturbations on \( x \). \( y \) is the standard output. The performance of the original prompt on the perturbed inputs is evaluated using a loss function $\mathcal{L}_{adv}$. The score is minimized when perturbed. After generating the perturbed sample prompt \( p \), we compute either its Levenshtein distance or semantic similarity \( ||x' - x|| < \epsilon \), depending on the nature of the perturbation. The perturbations are defined as follows: 
\begin{equation}
x' = x + \epsilon \cdot \arg\min \mathcal{L}_{adv}(x + g, y),
\label{equation:eq1}
\end{equation}
For both types of perturbation, we adopt an iterative interference approach. At each step, we select the sentence that affects the outcome while exhibiting the highest semantic or structural similarity to the original sentence. Our optimization objective is formulated as follows:
\begin{equation}
x' = x_0 + \mathcal{P}(x_0,g) + \mathcal{P}(x_1,g) + \ldots +  \mathcal{P}(x_{n-1},g),
\label{equation:eq2_}
\end{equation}
Where $\mathcal{P(\cdot)}$ denotes that the given input \( x \) is perturbed under the guide \( g \), \( g\) $\in$ \{\textit{P1}, \textit{P2}\}. \( x_n \) denotes the text generated under each iteration. After each iteration, we compute the Levenshtein distance or semantic similarity between all perturbed texts and their corresponding source texts, selecting the resulting text as input for the next iteration. However, such iterative perturbation is generally only effective for short texts, such as classification or simplification tasks. When applied to long-text inputs, such as text summarization, this approach faces two main issues: \textbf{\ding{182}} The model exhibits weak sensitivity to P1-type perturbations. \textbf{\ding{183}} The perturbation model for P2-type perturbations tends to produce shorter outputs, resulting in deviations from the original intent. To address these issues, for long-text inputs, we perturb only one sentence at a time. This sentence is randomly selected from the text in each iteration like \( s_i\) $\in$ $\{s_1, s_2, ..., s_n\}$. $s$ denotes each sentence in the long text. This strategy preserves the accuracy of the model’s response when perturbations are introduced and helps maintain the overall semantic relevance of the generated text.

\subsection{Results Analysis and Task Vulnerability}

\begin{figure}
    \centering
    \includegraphics[width=0.5\textwidth]{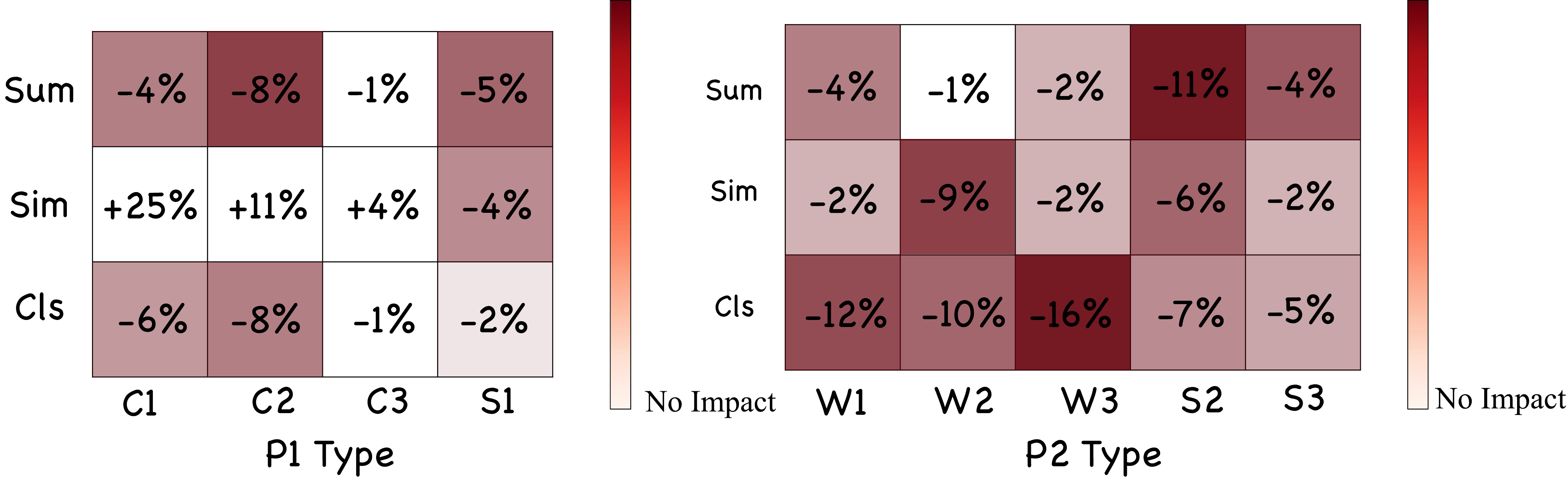}
    \caption{Heat maps showing the magnitude of the impact of each perturbation on different types of tasks, where darker colors indicate a stronger impact of the perturbation on this type of task, and vice versa a lower one. No color indicates no effect or a positive effect.}
    \label{fig:heat}
    \vspace{-10pt}
\end{figure}

To evaluate the effectiveness of our constructed dataset in generating prompt perturbations, we tested the performance of all prompts used in the Baseline model under the PertBench framework. The detailed results are presented in the experimental section, alongside those of our proposed method for comparison. To provide a clearer understanding of model vulnerabilities, we visualize the performance of each baseline across all perturbations using radar plots in Figure.~\ref{fig1:radar}. These plots reflect the outcomes on three tasks: the first three graphs correspond to different metrics for the text summarization task; the fourth graph presents results for the text simplification task; and the fifth graph shows results for the text classification task. From the visualizations, it is evident that most perturbations degrade task performance, indicating their effectiveness in identifying vulnerabilities. However, not all perturbations have a negative impact-some even improve performance. For instance, in the summarization task, perturbations C3 and W2 do not reduce model performance. In the simplification task, perturbations C1, C2, and C3 show no negative effect. Similarly, for the classification task, C3 and S1 yield positive effects in many cases. To further illustrate these findings, we include heatmaps showing the effects of various perturbations on manual prompts in Figure~\ref{fig:heat}. In these heatmaps, the numerical values represent the proportion of change compared to the original (unperturbed) input. The results clearly show that certain perturbations, particularly C3, have minimal or no negative impact across all three tasks. This suggests that the prompts involved in C3 are particularly robust to perturbations. We hypothesize that this is because the presence of input noise encourages the LLM to focus more on the core meaning, thereby aligning better with the objective of simplification.

\begin{figure*}[h]
    \centering

    \includegraphics[width=1\textwidth]{fig/workflow1.pdf}
    \caption{The workflow of an iteration of PGO. We present the template of PGO in~\autoref{temp}.}
    \label{fig1:pipeine}
\end{figure*}

\begin{table}
\centering
\resizebox{0.45\textwidth}{!}{
\begin{tabular}{c|c|c|c}
\hline
Perturbation & Summarization & Simplification & Classification \\
\hline
C1 & \ding{51} & \ding{55} & \ding{51} \\
C2 & \ding{51} & \ding{55} & \ding{51} \\
C3 & \ding{55} & \ding{55} & \ding{55} \\
W1 & \ding{51} & \ding{51} & \ding{51} \\
W2 & \ding{55} & \ding{51} & \ding{51} \\
W3 & \ding{51} & \ding{51} & \ding{51} \\
S1 & \ding{51} & \ding{51} & \ding{55} \\
S2 & \ding{51} & \ding{51} & \ding{51} \\
S3 & \ding{51} & \ding{51} & \ding{51} \\
\hline
\end{tabular}}
\caption{Sensitivity of each task on different perturbations, \ding{51} for sensitive and \ding{55} for robust.}
\label{Tab:com}
\end{table}

Specifically, we present the robustness relationship between each task and each perturbation. The detailed results are shown in the Table~\ref{Tab:com}. In the subsequent construction of robust prompts, we focus only on perturbations that lead to performance degradation in LLMs.

\section{Prompt Generation Framework}
To enhance auto-prompt robustness under perturbations, we propose \textbf{PGO}, a real gradient-free optimization framework. Unlike traditional methods requiring gradient access, PGO leverages pseudo guidance signals to iteratively generate and optimize prompts, improving robustness in black-box LLMs settings.

\subsection{Analyzing General Method}
To enhance prompt robustness for task performance, many researchers have turned to data augmentation, adding perturbed text to the training data and training the model on this augmented dataset to produce robust prompts. We conduct an ablation study on dataset with perturbed data.The details are in Experiments. The result shows, in some tasks, the performance of these prompts even declines, likely due to the excessive diversity of perturbation types introduced during augmentation. We contend that while basic data augmentation methods can be beneficial in certain contexts, they may inadvertently introduce noise that interferes with effective prompt generation and limits generalization to unseen perturbations.



\subsection{Framework of PGO}
PGO draws on the idea of adversarial training. We divide the process of optimization in two phases and uses a pseudo gradient to substitute the real gradient to add perturbation and optimization. It first iteratively adds perturbations within the specified range to undisturbed inputs by adjusting them along the gradient direction to maximize disruption. The perturbed samples generated are then used for training based on a defined loss function. The workflow of our method is shown in Figure~\ref{fig1:pipeine}. PGO contains many iterations and every iterations include two key components:
\begin{itemize}[left=0pt]
\item \textbf{Pseudo Gradient Perturbation Phase}. PGO begins with manually crafted prompts and unperturbed text as the initial input. In Section.2, we categorize nine distinct perturbations into two types and design tailored perturbation methods for each category. These methods generate perturbed samples that simulate various perturbation scenarios, allowing the subsequent optimization algorithms to build robustness across all types of perturbations. This approach ensures that the model remains effective when faced with diverse perturbations.
\item \textbf{Pseudo Gradient Optimization Phase}. At this stage, we employ an iterative optimization method that incorporates a "gradient" mechanism to guide the LLMs in refining the prompts. For each generated prompts, we select the candidate with the best performance on the validation set and retain it for the next iteration.
\end{itemize}

\subsection{Pseudo Gradient Perturbation Phase}
At this stage, we guide the construction of the perturbation dataset through the perturbation category, and maintain the similarity as the purpose of building the benchmark. However, for the two distinct perturbation types, \textit{P1} and \textit{P2}, we design two different modes of perturbation. This ensures that the text used in the optimization process encompasses all perturbations within each type. As a result, the generated prompts demonstrate robustness against all attacks corresponding to their respective types. For the \textit{P1} type, we propose a \textbf{mix-mode}, which is implemented as follows:
\begin{equation}
x' = x_0 + \mathcal{P}(x_0,g_1) + \mathcal{P}(x_1,g_2) + \ldots +  \mathcal{P}(x_{n-1},g_n),
\label{equation:eq2}
\end{equation}
Where $\mathcal{P(\cdot)}$ denotes that the given input \( x \) is perturbed under the guide \( g \), \( g_n \) $\in$ \textit{P1}, \( x_n \) denotes the text generated under each perturbation. After each perturbation, we compute the Levenshtein distance between all perturbed texts and their corresponding source texts, selecting the resulting text as input for the next iteration. The rationale behind this design is that, under the perturbation of the \textit{P1} mode, the superposition of each perturbation has minimal impact on the effect of other perturbations. 
This design enables to generate results that effectively incorporate all types of perturbations. 

In the perturbation of \textit{P2} mode, we propose a \textbf{combined-mode} as follows:
\begin{equation}
\mathcal{U}(x') = \mathcal{P}(x,g_1) \cup \mathcal{P}(x,g_2) \cup \ldots \cup \mathcal{P}(x_,g_n),
\label{equation:eq3}
\end{equation}
where $\mathcal{U(\cdot)}$ denote the set generated after all perturbations. Unlike \textit{P1} type perturbations, \textit{P2} perturbations involve semantic modifications through operations such as rewriting, changing grammar and so on.  To ensure that the meaning remains aligned with the original sentence, thus preventing any adverse impact on the performance of the LLMs.  We rely on semantic similarity as the evaluation criterion. We choose to combine and input them into the pseudo gradient optimization phase.

\begin{table*}[!htbp]
\centering
\small
\begin{tabular}{c|ccc|ccc|c|c}
\toprule
\multirow{2}{*}{Methods} &
  \multicolumn{3}{c|}{\textbf{P1 perturbation}} &
  \multicolumn{3}{c|}{\textbf{P2 perturbation}} &
  \textbf{P1 perturbation} & \textbf{P2 perturbation}  \\
  \cmidrule{2-9}
 & Rouge-1 & Rouge-2 & Rouge-L &  Rouge-1 & Rouge-2 & Rouge-L & SARI & SARI \\ \midrule
DA & 11.39 & 2.50 & 9.61 & 12.19 & 2.84 & 10.39 & 44.87 & 44.28 \\
MI & 14.98 & 3.04 & 12.59 & 15.00 & 3.03 & 12.68 & 36.67 & 36.51 \\
EvoPrompt & 17.52 & 3.14 & 14.97 & 17.19 & 2.93 & 14.69 & 44.61 & 46.56 \\
PGO$^{*}$ & 18.20 & 2.93 & 15.37 & 18.22 & 3.12 & 15.64 & 45.28 &  45.73\\
\textbf{PGO} & \textbf{21.25} & \textbf{4.44} & \textbf{16.66} & \textbf{21.90} & \textbf{4.73} & \textbf{16.81} & \textbf{45.79} & \textbf{47.85} \\ 
\bottomrule  
\end{tabular}
\caption{The average Rouge-1($\uparrow$) score, Rouge-2($\uparrow$) score and Rouge-L($\uparrow$) score on the text summarization task and average SARI($\uparrow$) scores on the text simplification task for our method and the comparison methods.}
\label{table2:lan_ge}
\end{table*}

\subsection{Pseudo Gradient Optimization Phase}
In the optimization stage, we utilize the perturbed sample \( x' \) generated in the first stage. We then analyze the differences between \( x' \) and the corresponding original text \( x \), and use these differences to compute the gradient \( g' \) when the guide is used as the optimization prompt. The detailed process is as follows:
\begin{equation}
g' =  \mathcal{G}(\mathcal{D}(x_0,x_0^{'}) \cup \mathcal{D}(x_1,x_1^{'})  \cup  \ldots \cup \mathcal{D}(x_n,x_n^{'})),
\label{equation:eq4}
\end{equation}
Where $\mathcal{D}(\cdot)$ denotes the generating difference and $\mathcal{G}(\cdot)$ denotes the generation of  general guidance. In the iterative optimization process following gradient generation, we also use the prompt's score on the task as the loss function. However, unlike the previous stage, we select the prompt with the highest score at each iteration to maximize task performance. To ensure that the final prompt is robust to all perturbations, we calculate the score for each perturbation across the vulnerabilities of the target task. The specific loss function is as follows:
\begin{align}
\mathcal{L}_{opt}(p) = \mathbb{E}_{(x', y) \sim D} \big[ 
    & \mathcal{L}(x_1^{'}, y; p) + \ldots \notag \\
    & + \mathcal{L}(x_n^{'}, y; p) 
\big].
\label{equation:eq5}
\end{align}
Where $\mathcal{L}_{opt}(\cdot)$ represents the optimization loss. $\mathcal{L}(\cdot)$ denotes the loss per class of perturbation. In summary, the formula for each round of prompt optimization is $p' = p + \nabla_{p} \mathcal{L}_{opt}(p).$
In prompt generation, to expand the range of options, we rewrite the generated prompts at each iteration, increasing the likelihood of discovering better alternatives. During the intermediate optimization iterations, we consistently select the optimal prompt from each round to proceed to the next iteration of the loop.

\section{Experiments}
\subsection{Implementation Details and Baselines}
In the experiments, we use GPT-3.5-turbo to do the optimization training and use \textbf{GPT-3.5-turbo}, \textbf{GPT-4o-mini} and \textbf{Llama2-7b} to test the effectiveness of the instructions generated PGO. For perturbations of type \textit{P1}, we select five examples in each iteration, while for perturbations of type \textit{P2}, we select three examples per iteration, considering the number of perturbed examples generated. After five iterations, we choose the prompt with the highest score on the training set and evaluate its performance on the test set. More experiment settings in the~\autoref{set}. In the evaluation, we compare the prompts generated by PGO with the following methods:
\begin{itemize}[left=0pt]
\vspace{-3pt}
\item \textbf{Manual Instructions(MI)} and \textbf{Natural Instructions(NI)}~\cite{mishra2021cross}: MI are based on existing work that is task-specific guidelines. Including classification task~\citet{zhang2023sentiment}, text simplification task~\citet{zhang2023multi} and summarization task~\citet{sanh2021multitask}. NI contains manually designed prompts across a  diverse range of datasets and tasks. 
\item \textbf{APE}~\cite{zhou2022large}, \textbf{InstructZero}~\cite{chen2023instructzero}, \textbf{INSTINCT}~\cite{lin2004rouge}: These papers leverages the reasoning ability of LLMs to generate prompts based on both the inputs and the corresponding answers.
\item \textbf{EvoPrompt}~\cite{guo2023connecting}: Evoprompt uses genetic algorithms to optimize prompts for classification and generation tasks.

\item \textbf{Data Augmentation(DA)} and \textbf{PGO$^{*}$}: DA uses the data augmentation method as baseline, which takes the perturbed text as input data and iteratively optimizes it, to explore whether this method can remain robust to all types of text perturbations. PGO$^{*}$ uses the simple iterative method to optimize the prompts.
\end{itemize}
We also analysis the performance of adding suggestive suffix without PGO directly in~\autoref{add_suffix} and the cost of PGO in~\autoref{cost}. Our optimal results are shown in~\autoref{prompts}.

\subsubsection{Metrics}
In the \textbf{classification tasks}, we use the prediction accuracy as the score. In \textbf{language generation tasks}, we use SARI~\cite{xu2016optimizing} as the metrics which is an n-gram-based scoring system extensively utilized for text editing tasks. For the \textbf{text summarization task}, we use Rouge-1, Rouge-2, Rouge-L as metrics~\cite{lin2004rouge}, which widely used to evaluate the quality of generated text tasks. They focus on the number of n-grams with the same outcome and the overlap of the longest common subsequence.


\begin{table*}
\centering
\resizebox{\textwidth}{!}{
\begin{tabular}{c|cccccc|cccccc||c}
\toprule
\multirow{2}{*}{Methods} &
  \multicolumn{6}{c|}{\textbf{P1 perturbation}} &
  \multicolumn{6}{c||}{\textbf{P2 perturbation}} &
  \multirow{2}{*}{\textbf{Avg.}} \\
  \cmidrule{2-13}
   & \textbf{CR} & \textbf{SST-2} & \textbf{MR} & \textbf{SST-5} & \textbf{TREC} & \textbf{AG} & \textbf{CR} & \textbf{SST-2} & \textbf{MR} & \textbf{SST-5} & \textbf{TREC} & \textbf{AG} \\  \midrule
 DA & 88.5 & 89.8 & 85.5 & 44.0 & 43.8 & 86.3 & 85.1 & 85.5 & 81.2 & 40.1 & 50.3 & 86.7 & 72.3 \\
 MI & 89.3 & 88.8 & 85.5 & 45.5 & 50.3 & 87.0  & 85.0 & 85.5 & 82.0 & 43.6 & 46.2 & 86.2 & 72.9 \\
  NI & 83.3 & 80.3 & 80.0 & 23.3 & 45.3 & 65.3 & 79.1 & 80.0 & 77.0 & 20.0 & 43.4 & 61.8  & 61.6\\
EvoPrompt & 63.8 & 80.5 & 78.1 & 10.5 & 1.5$^?$ & 67.3 & 62.8 & 78.1 & 80.6 & 8.6 & 1.5$^?$ & 66.8 & 50.4\\
  PGO$^{*}$ & 85.5 & 86.0 & 81.0 & 45.3 & 49.8 & 85.5 & 80.6 & 81.5 & 78.1 & 40.9 & 49.9 & 84.3 & 70.8\\
  \textbf{PGO} & \textbf{89.8} & \textbf{90.0} & \textbf{86.3} & \textbf{46.0} & \textbf{52.3} & \textbf{88.5} & \textbf{85.7} & \textbf{86.3} & \textbf{82.8} & \textbf{44.9} & \textbf{51.6} & \textbf{87.7} &\textbf{73.2} \\ 
\bottomrule  
\end{tabular}
}
\caption{Average score($\uparrow$) of the prompts from different method on six classification datasets. $?$ Indicates that we provide specific analysis of the value of this indicator in~\autoref{ana}. AG indicates AG'news dataset.}
\label{table:cls}
\vspace{-10pt}
\end{table*}

\subsection{Data Augmentation}
To evaluate the limitations of the data augmentation method, we exclude the use of PGO during testing. Instead, we treat all the perturbed data as the training set and allow the LLMs to generate prompt based on this training set. The results for three different tasks are presented in the \textit{DA} row across the Table.~\ref{table2:lan_ge}, Table.~\ref{table:cls}. As shown, the prompts generated using the data augmentation method lack robustness to perturbations across all three tasks. Furthermore, their performance is even worse than unoptimized prompts when tested on handwritten perturbations. We attribute this degradation to the excessive diversity of perturbations, which hinders the LLMs's ability to focus accurately on the perturbations, leading to a decline in performance.

\subsection{Experiments Results}

In this part, we use GPT-3.5-turbo to test the performance of prompts generated by PGO and other comparison methods on the three tasks. The results of \textbf{text summarization} task and \textbf{text simplification} task are shown in the Table~\ref{table2:lan_ge}. Compared to previous prompts and their generation methods, the prompts generated by PGO demonstrate a significant performance improvement on the perturbed datasets.  It is worth noting that in the text summarization task, the model outperforms the second-best baseline by an average of 20\%, 46\%, and 7.5\% in ROUGE-1, ROUGE-2, and ROUGE-L scores, respectively. For \textbf{classification} task, the results for the perturbations from class P1 and class P2 are presented in the Table~\ref{table:cls}. The table demonstrates that the prompts generated by PGO outperform existing prompts across the six text understanding datasets, exhibiting notable stability against the seven perturbations to which PGO itself is not inherently robust. It is worth mentioning that PGO achieves a 4\% improvement on TREC under the P1-class perturbation and a 3\% improvement on SST-5 under the P2-class perturbation. In addition, we observe that the cue words generated by our method demonstrate notably stronger retention robustness on multiple-choice datasets such as SST-5, TREC, and AG News. We note that EvoPrompt performs significantly worse than our method on the SST-5 and TREC datasets, and We attribute this to deficiencies in the generated prompts resulting from its prompt optimization method; a detailed analysis can be found in the~\autoref{ana}. For methods APE~\cite{zhou2022large} InstructZero~\cite{chen2023instructzero}, INSTINCT~\cite{lin2004rouge}, PGO still maintains the optimal performance on all tasks, and more specific results are in the~\autoref{APE}.

\subsection{Transferability and Generalization}

\subsubsection{Model Transferability}
To evaluate the transferability of our generated prompts, we select the text summarization task and assess the effectiveness of our method on \textbf{GPT-4o-mini} and \textbf{LLaMA2-7B}. The results demonstrate that, compared to other methods, our approach consistently maintains strong performance across different models. Additional transferability experiments are provided in the~\autoref{trans}.

\subsubsection{Perturbation Generalization}
In this section, we evaluate the transferability of prompts trained under different perturbation types. Specifically, we adopt a cross-testing strategy: prompts trained under perturbation P1 are tested on robustness against perturbation P2, and vice versa. The results are presented in the Table.~\ref{tab:pgpt}. Notably, our PGO$_{change}$ performs only slightly worse than PGO across most metrics, indicating that prompts optimized for one class of perturbations can still exhibit strong robustness when applied to other perturbation types.

\begin{table}
\footnotesize
\resizebox{0.5\textwidth}{!}{
\begin{tabular}{c|c|c|c|c}
\toprule
\textbf{Model} & \textbf{Method} & \textbf{Rouge-1} & \textbf{Rouge-2} & \textbf{Rouge-L} \\ \midrule
\multirow{3}{*}{GPT-4o-mini} & MI & 16.10 & 2.84 & 13.00 \\
& EvoPrompt & 16.83 & 2.63 & 14.33 \\
& \textbf{PGO} & \textbf{19.49} & \textbf{3.63} & \textbf{14.97} \\ \midrule
\multirow{3}{*}{LLaMA2-7b} & MI & 16.45 & 2.95 & 13.80 \\ 
& EvoPrompt & 14.80 & 2.56 & 12.85 \\
& \textbf{PGO} & \textbf{18.30} & \textbf{3.64} & \textbf{14.71} \\ 
\bottomrule
\end{tabular}}
\caption{Transferability of prompts across methods and models in text summarization.}
\label{tab:trans}
\vspace{-15pt}
\end{table}
\subsection{Ablation Study}

\subsubsection{The Effectiveness of Pseudo Gradient}
To evaluate the effectiveness of first stage in the prompt optimization, we removed the pseudo gradient perturbation part from PGO and used only a simple iterative optimization method. This allowed us to isolate and assess the impact of first phase. The results are show in \textit{PGO$^{*}$} rows in the Table~\ref{table2:lan_ge} and Table.~\ref{table:cls}. The experimental results demonstrate that, across all tasks, prompts generated purely through iterative optimization perform worse than those generated by PGO across all types of perturbations. This proves the effectiveness of our pseudo gradient strategy. 

\subsubsection{Iteration Numbers}
\begin{figure}
    \centering
    \includegraphics[width=0.5\textwidth]{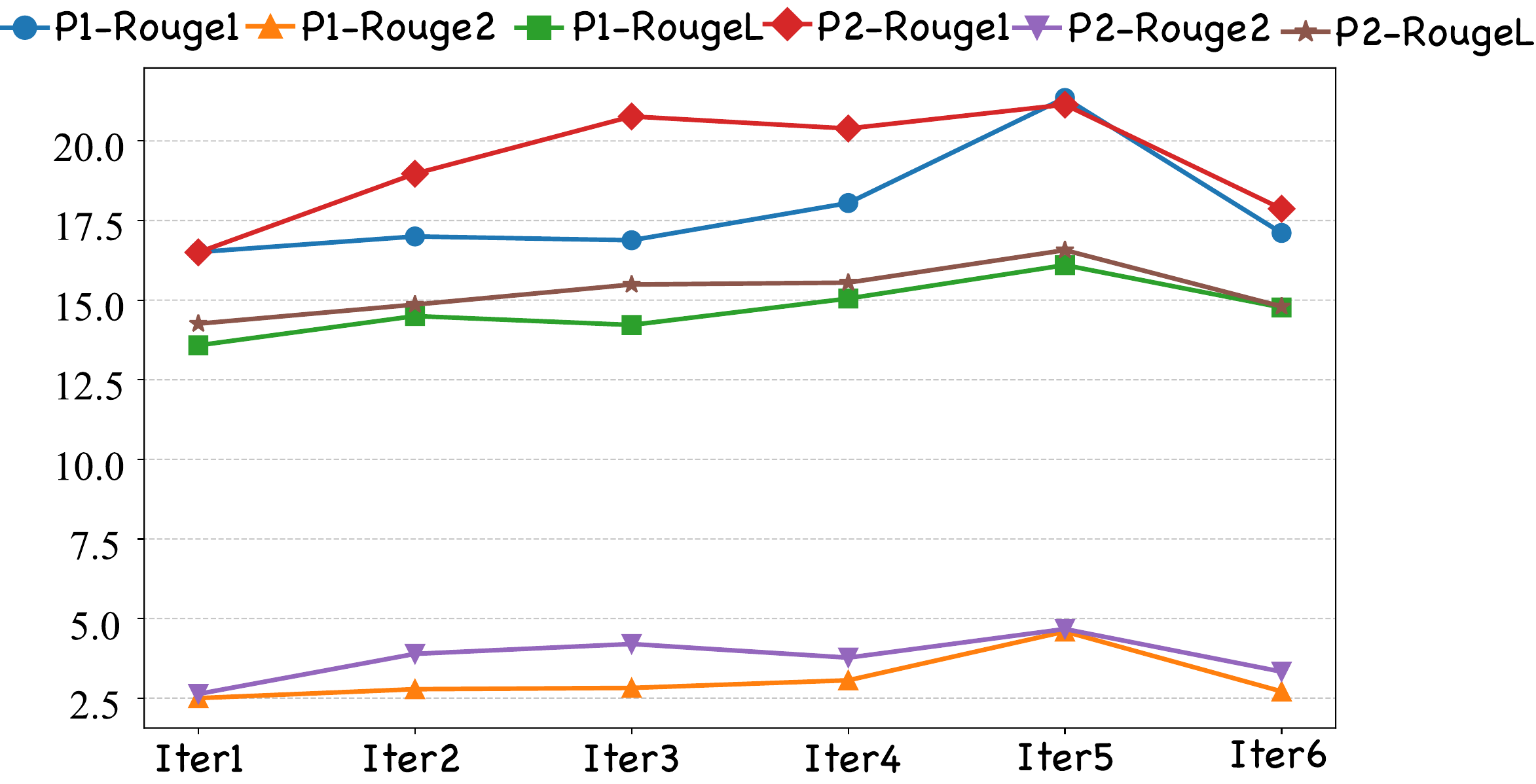}
    \caption{Relationship between iteration count and PGO prompt performance.}
    \label{fig:iter}
    \vspace{-10pt}
\end{figure}

In our optimization process, we set the number of iterations to five, as we observed that the prompts optimized by the LLMs achieved optimal performance at the fifth iteration. To illustrate this, we used the text summarization task as an example. We tested the performance of prompts generated across six iterations (from 1 to 6), and the results are presented in the Figure.~\ref{fig:iter}.
For each iteration, we evaluate the prompts across different perturbation types and calculate their average performance. The results show a consistent improvement in prompt performance during the initial rounds, with the optimal performance observed around the fourth or fifth iteration. This trend suggests that the LLMs effectively refines the prompts through iterative gradient-guided optimization and semantic space exploration, progressively approaching the optimal solution. However, beyond the fifth iteration, the performance of the prompts have a slightly declines. In order to speed up the inference experiment and reduce the cost, we choose the number of iterations to be 5. We present more discussion in~\autoref{dis}.
\begin{table}
\footnotesize
\resizebox{0.5\textwidth}{!}{
\begin{tabular}{c|c|c|c|c}
\toprule
\textbf{Type} & \textbf{Method} & \textbf{Rouge-1} & \textbf{Rouge-2} & \textbf{Rouge-L} \\ \midrule
\multirow{2}{*}{P1} 
& PGO$_{change}$ & \textbf{21.35} & \textbf{4.59} & 16.10 \\
& PGO & 21.25 & 4.44 & \textbf{16.66} \\ \midrule
\multirow{2}{*}{P2} &  PGO$_{change}$ & 21.26 & 4.31 & 16.33 \\
& PGO & \textbf{21.90} & \textbf{4.73} & \textbf{16.81} \\ 
\bottomrule
\end{tabular}}
\caption{The average of Rouge-1($\uparrow$), Rouge-2($\uparrow$) and Rouge-L($\uparrow$) of prompts trained by different perturbation on the text summarization task.PGO$_{change}$ means test the results on other perturbation.}
\label{tab:pgpt}
\end{table}
\section{Related Work}

\subsection{Prompt Optimization}
Manually crafted prompts often fail to improve LLM performance, motivating the rise of prompt optimization. ~\citet{li2023tuna} introduces a fine-tuning approach based on context and probability ordering, while reinforcement learning methods such as Eureka~\cite{ma2023eureka}, Prompt-OIRL~\cite{sun2023query}, and RLPrompt~\cite{deng2022rlprompt} further enhance optimization. Recent work leverages LLMs themselves for prompt generation and refinement, including APE~\cite{zhou2022large}, self-correction via pseudo gradients~\cite{pryzant2023automatic}, genetic algorithms~\cite{guo2023connecting}, and robustness studies~\cite{jin-etal-2024-impact,zhou2024robust}. However, most methods rely on clean data, neglecting natural input perturbations that degrade performance. Our approach addresses this by emphasizing robustness against such perturbed inputs.

\subsection{Improving the Robustness of LLMs}
Recent studies explore adversarial attacks on LLMs, where slight input perturbations (e.g., spelling errors) mislead models into incorrect outputs. ~\citet{zhou2024mathattack} target math reasoning, while ~\citet{zhu2023promptbench} categorize hint attacks into four types, detailed by ~\citet{xu2024an}. Common approaches include word-level~\cite{wang2023large, zhou2024evaluating} and sentence-level~\cite{gu2023towards, dong2021sentence} attacks. ~\citet{miao2024autonomous} introduce attacks for T2M, and ~\citet{raina2024llm} reveal Judge-LLMs’ vulnerability, inflating evaluation scores. Other works~\cite{yao2023llm, kumar2023certifying} study adversarial behaviors and propose defense strategies, including adversarial hallucination~\cite{kumar2023certifying} and refusal-based defenses~\cite{sheshadri2024targeted, xhonneux2024efficient, lin2024large}. Adversarial training methods like ReFAT~\cite{sheshadri2024targeted}, Latent Adversarial Training~\cite{xhonneux2024efficient}, and LLAMOS~\cite{lin2024large} further enhance LLM robustness against attacks. 

\section{Conclusion}
In this paper, we highlight the vulnerability of existing automatic prompt word generation methods to input perturbations. To systematically evaluate this limitation, we introduce \textbf{PertBench}, a benchmark dataset that spans multiple tasks and incorporates a diverse set of perturbation types at the character, word, and sentence levels. PertBench enables a comprehensive assessment of prompt robustness under a wide range of input disturbances. Furthermore, we propose a novel prompt optimization method, \textbf{PGO}, that treats perturbation types as pseudo-gradient signals to guide the generation of more robust prompts in a gradient-free manner. Extensive experiments across various tasks and perturbation settings demonstrate the effectiveness and superior robustness of our approach compared to existing methods. 


\section{Limitation}
We employed an efficient, training-free pseudo-gradient method. In the future, additional gradient-free techniques could be incorporated into our framework to further reduce generation costs and enhance performance. Moreover, although our method significantly outperforms existing approaches, its reliance on black-box models may limit the explainability of its superiority. Therefore, more advanced explainable AI (XAI) techniques may be required to better interpret this phenomenon.

\bibliography{custom}

\appendix

\section{Details of Benchmark}
\label{bench_detail}

\begin{table}
\centering
\renewcommand{\arraystretch}{1.25}
\resizebox{0.5\textwidth}{!}{
\begin{tabular}{l||ll}
\toprule
\textbf{Dataset} & \textbf{Task} & \textbf{Number}\\
\midrule
SST-2 &\{positive, negative\} & 1200 \\
CR & \{positive, negative\} & 1200 \\
MR & \{positive, negative\} & 1200 \\
SST-5 & \{terrible, bad, okay, good, great\} & 1200\\
AG’s News & \{World, Sports, Business, Tech\} & 1200 \\
TREC & \{Description, Entity, Expression, Human, Location, Number\}  & 1200\\
ASSET & Text Simplification   & 987 \\
XSUM & Text Summarization  & 720 \\
\bottomrule
\end{tabular}
}
\caption{Statistics of our generating datasets for classification task and language generation task used in this work. }
\label{tab:stat}
\end{table}

The Table.~\ref{tab:stat} shows the statistics of datasets we made for classification, text simplification, and text summarization tasks under different perturbations. Each dataset contains multiple sub-datasets, the total amount of generated data includes only perturbations to which the model is not robust for the given task. For example, the XSum dataset contains 7 sub-datasets corresponding to C2, C1, S1, W3, S3, S2, and W1 perturbations. For perturbations where the model is already robust, we constructed only 200 test samples per task. The robustness of the perturbation category is in the previous section.

\subsubsection{Similarity between PertBench and Raw Data}
\begin{table}
\footnotesize
\large
\resizebox{0.45\textwidth}{!}{
\begin{tabular}{c|c|c}
\toprule
Datasets & Levenshtein distance & Semantic similarity  \\ \midrule
XSum  & 98.42 & 98.31  \\
Asset  & 89.55 & 91.17 \\
CR & 93.38  & 85.92  \\
SST-5  & 91.22 & 79.75    \\ 
AG'News & 96.63 & 83.91   \\ 
MR & 92.42  & 82.33  \\ 
TREC & 91.41  &81.44 \\
SST-2 & 91.96  & 80.86 \\
 \bottomrule
\end{tabular}}
\caption{The averagesemantic similarity and Levenshtein distance before and after attack in 3 tasks, 8 datasets.}
\label{tab:att}
\end{table}

In this part, we measure the similarity score between the generated benchmark and the original data set to prove that our dataset retains a high similarity with the original dataset. From the Table.~\ref{tab:att}, we can observe that for long text inputs, such as those in the XSum dataset, the Levenshtein distance and semantic similarity between the original and perturbed text remain above 98\% for both types of perturbations (P1 and P2). In contrast, for shorter text inputs, such as those in the Asset dataset and the six classification tasks, even small changes can significantly impact the similarity measures. However, our experiments show that the Levenshtein distance similarity for P1 perturbed data remains above 90\%, while the semantic similarity for P2 perturbed data stays above 80\%, except for the SST-5. This indicates that our attack effectively preserves the essential information of the original sentence while introducing controlled perturbations.

\section{Experiments Setting}

\subsection{Hpyper Parameters}
\label{set}
Our PGO algorithm is based on GPT-3.5-turbo for generation, with a total of 5 optimization iterations. During the perturbation phase, we set the number of iterative attacks to 3. In the combined perturbations and prompt optimization stages, we configured GPT-3.5-turbo with a Top-p value of 0.95 and a temperature of 1 to ensure both the robustness of the perturbations and the diversity of the generated outputs. For the testing phase, we set Top-p to 1 and temperature to 0, ensuring that the model produces consistent, fixed outputs.

\subsection{Template}
\label{temp}
\begin{figure}
    \centering
    \includegraphics[width=0.45\textwidth]{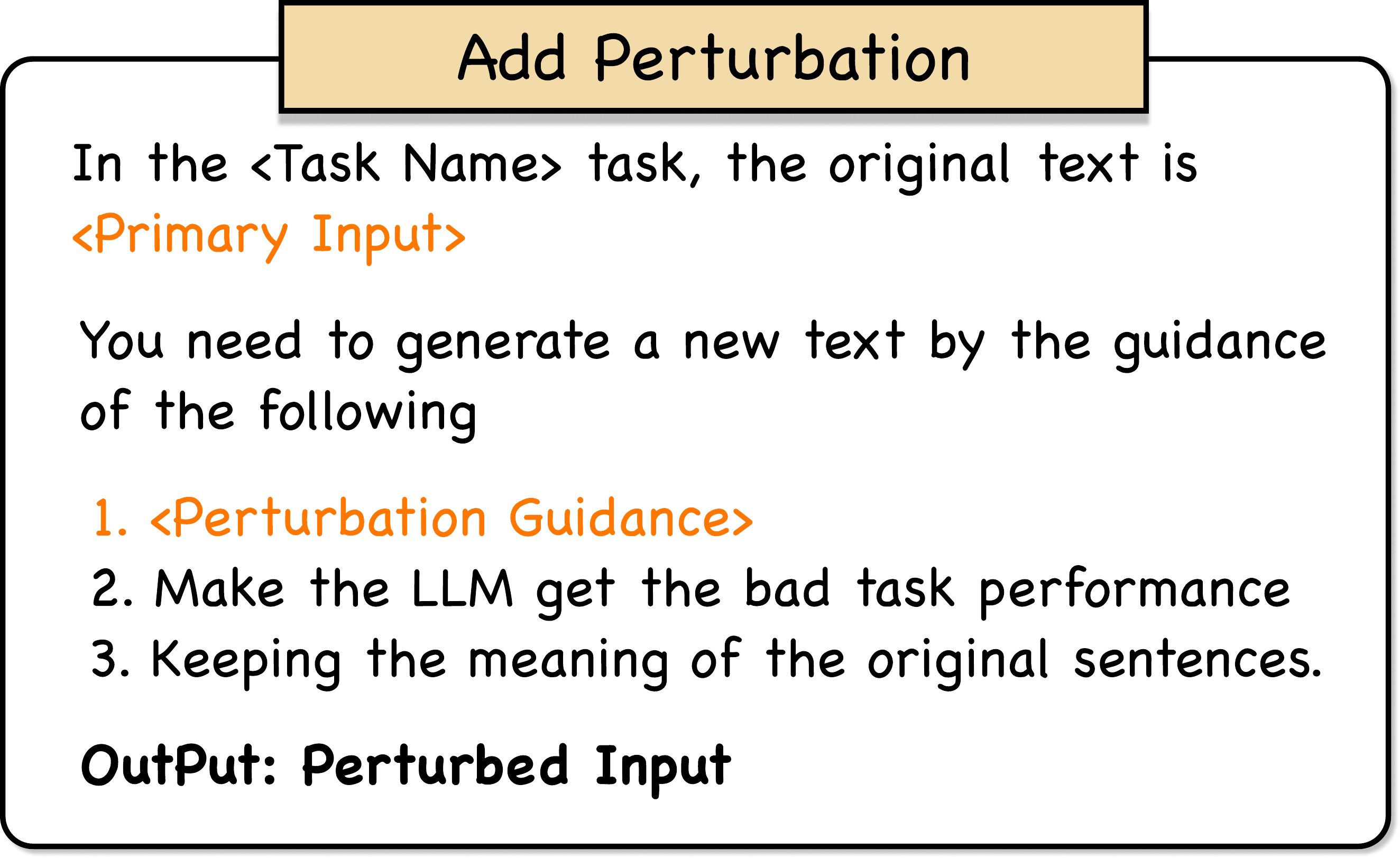}
    \caption{The template of generating the optimization gradient.}
    \label{fig:tem0}
\end{figure}

\begin{figure}
    \centering
    \includegraphics[width=0.45\textwidth]{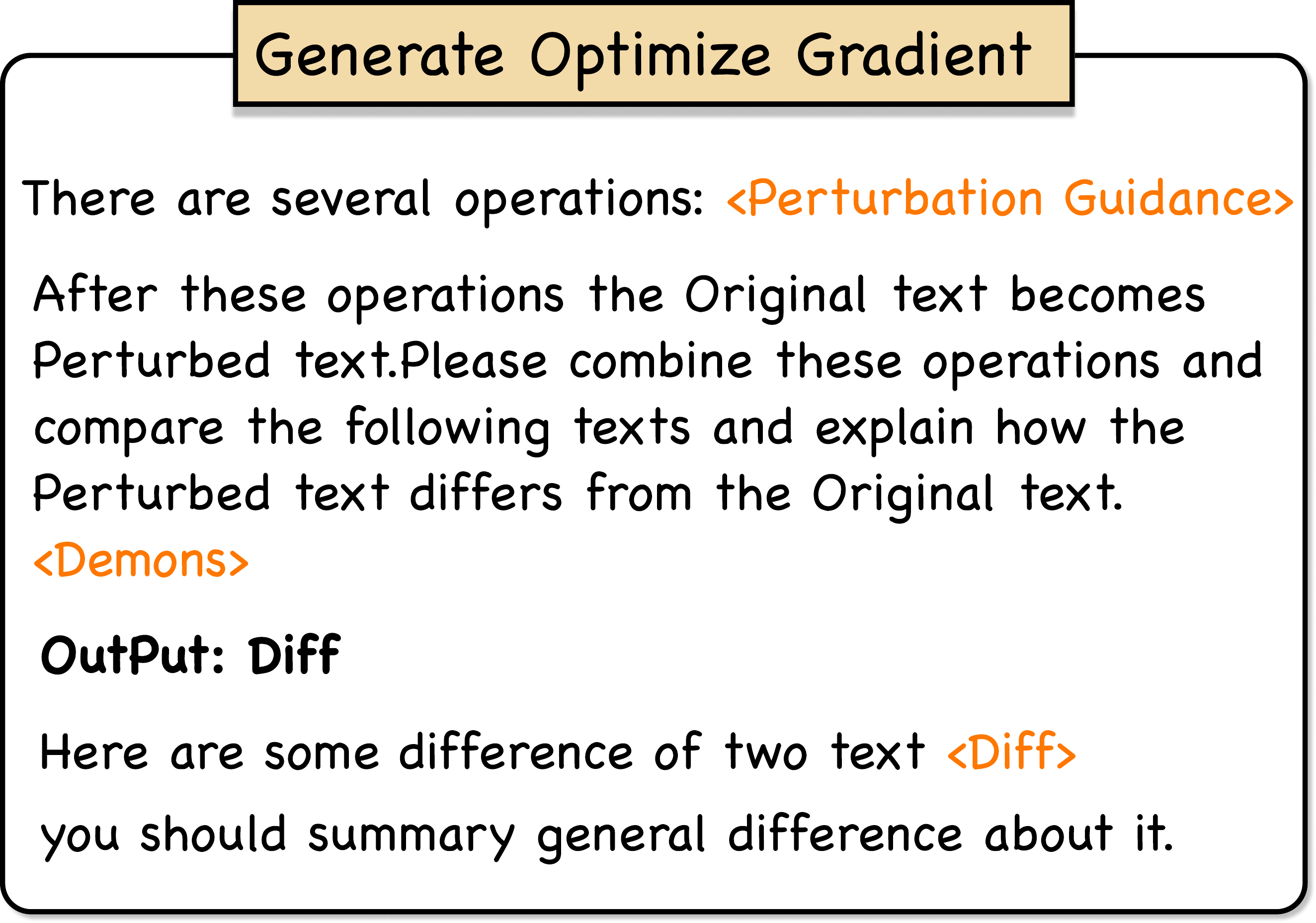}
    \caption{The template of generating the optimization gradient.}
    \label{fig:tem1}
\end{figure}

\begin{figure}[!ht]
    \centering
    \includegraphics[width=0.45\textwidth]{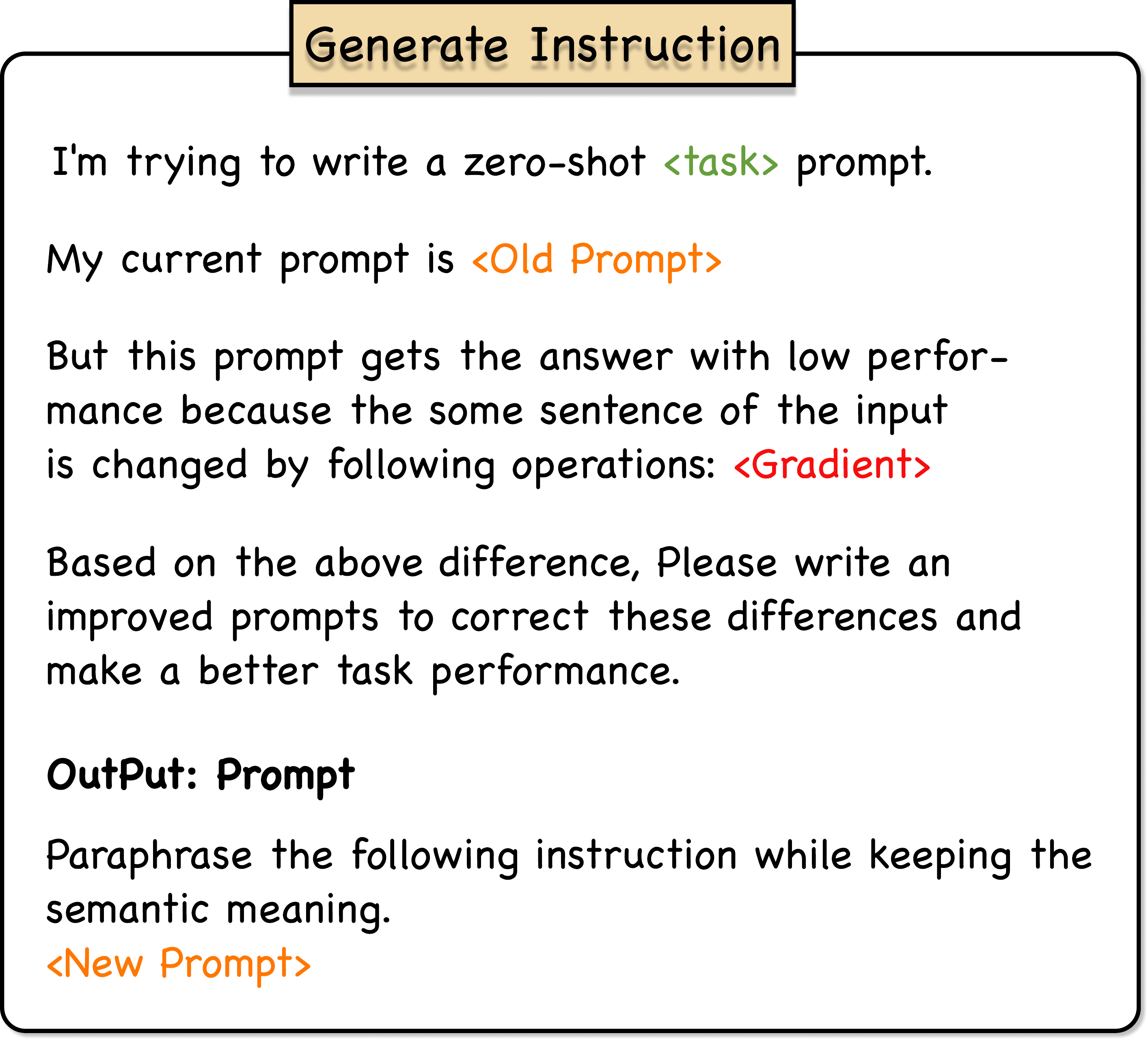}
    \caption{The template of using gradient to generate new instructions and paraphrase them.}
    \label{fig:tem2}
\end{figure}

In this section, we give a complete template for the perturbation adding phase. Figure.~\ref{fig:tem0}, the prompt optimization phase. Figure.~\ref{fig:tem1} provides a detailed illustration of the gradients generated during the optimization phase. Figure.~\ref{fig:tem2} offers a comprehensive explanation of the prompt generation process, where gradients guide the LLMs, and highlights how prompt richness is enhanced through the rewriting process. The details of the specific implementation in LLMs are shown in Figure.~\ref{fig:details}.

\begin{figure*}[h]
    \centering
    \includegraphics[width=\textwidth]{fig/details.pdf}
    \caption{PGO is based on the details of the LLMs implementation, with the left half representing the pseudo gradient perturbation phase and the right half representing the pseudo gradient optimization phase, where orange represents the input text, blue represents the gradient, red represents the prompts}
    \label{fig:details}
\vspace{-10pt}
\end{figure*}

\section{Additional Results}
\label{more}
\subsection{Experimental results for more methods}
\label{APE}
We compare our method against additional baselines (APE, InstructZero and INSTINCT) using GPT-4o-mini, and the results on both text generation and text understanding tasks are presented in the Table.~\ref{table2:sum_p2_APE}, Table.~\ref{table:sim_ape} and Table.~\ref{table:cls_ape}. It can be seen that our method has strong robustness on perturbed data and performs well in different tasks.

From the experimental results, we observe that the three baseline methods are not robust to perturbed text, with their performance consistently degrading after perturbations are introduced. Notably, the performance drop is more pronounced on multiple choice tasks such as TREC and SST-5. We believe this is because these methods do not explicitly provide an initial prompt, but instead attempt to infer it from the input and output. In the context of multiple choice tasks, this inference-based approach may struggle to effectively capture all options and accurately identify the target category, which likely contributes to the lower performance.

\subsection{The stability of the results}
To evaluate the performance stability of all the generated prompts, we selected the top five prompts and calculated the variance of their corresponding metrics, as well as the ratio of the variance to the mean. The results are presented in Table.~\ref{tab:ratio_sum}.

From the results, we observe that for both summarization and simplification tasks, the standard deviation of ROUGE-1, ROUGE-2, ROUGE-L, and SARI scores for our generated prompts is consistently below 1.0, with most tasks exhibiting values under 0.5. Furthermore, the standard deviation-to-average ratio remains under 10\%, indicating the stability of our prompts in evaluating text similarity. For classification tasks, we evaluated three datasets including: CR, SST-5 selected from the six datasets used in the paper. These datasets encompass binary classification, multi-class classification, sentiment classification, and topic classification. Our results show that the standard deviation-to-mean ratio remains below 10\%, further demonstrating the robustness and stability of our prompt across diverse classification tasks.

\begin{table*}
\centering
\small
\setlength{\tabcolsep}{0.4mm}
\begin{tabular}{c|ccc|ccc|ccc|ccc}
\toprule
\multirow{2}{*}{Methods} &
  \multicolumn{3}{c|}{\textbf{W3 perturbation}} &
  \multicolumn{3}{c|}{\textbf{W1 perturbation}} &
  \multicolumn{3}{c|}{\textbf{S2 perturbation}}  & 
  \multicolumn{3}{c}{\textbf{S3 perturbation}} \\
  \cmidrule{2-13}
 & Rouge-1 & Rouge-2 & Rouge-L &  Rouge-1 & Rouge-2 & Rouge-L &  Rouge-1 & Rouge-2 & Rouge-L &  Rouge-1 & Rouge-2 & Rouge-L \\ \midrule
APE & 17.52 & 3.94 & 14.15 & 17.67 & 3.70 & 14.44 &	17.42 &	3.69 & 14.09 & 18.46 & \textbf{4.42} &15.14\\
InstructZero & 17.47 & \textbf{4.62} & 14.27 & 16.45 & 3.69 & 13.84 &17.46 &\textbf{4.20} & 14.99 & 17.12 & 3.56 & 	14.03 \\
INSTINCT & 17.01 & 4.34 & 14.33 & 15.44 & 3.40 & 12.86 & 15.63 &3.18&	12.96 &	17.61  & 3.83 & 14.17 \\
\textbf{PGO} & \textbf{18.79} & 4.38 & \textbf{15.05} & \textbf{18.20} & \textbf{4.06} & \textbf{14.92} & \textbf{17.78} & 3.98 & \textbf{14.71} & \textbf{18.96} & 4.33 & \textbf{15.45}   \\ 
\bottomrule  
\end{tabular}
\caption{The Rouge-1($\uparrow$) score, Rouge-2($\uparrow$) score and Rouge-L($\uparrow$) score obtained by the prompts generated by the other methods for the task where the text summarization task is weak under the \textit{P2} class perturbation.}
\label{table2:sum_p2_APE}
\end{table*}
\begin{table*}
\large
\resizebox{\textwidth}{!}{
\begin{tabular}{c|cc|cc|cc|cc|cc|cc}
\toprule
\multirow{2}{*}{Type} &
  \multicolumn{2}{c|}{\textbf{Rouge-1}} &
  \multicolumn{2}{c|}{\textbf{Rouge-2}} &
  \multicolumn{2}{c|}{\textbf{Rouge-L}} &
  \multicolumn{2}{c|}{\textbf{SARI}} &
   \multicolumn{2}{c|}{\textbf{CR}} &
    \multicolumn{2}{c}{\textbf{SST-5}}\\
  \cmidrule{2-13}
 & R1-std & std/avg & R2-std &  std/avg & RL-std & std/avg & SARI-std & std/avg & CR-std &std/avg& SST5-std & std/avg \\ \midrule
P1 & 1.04 & 5.23\% & 0.23 & 6.37\% & 0.38 & 2.47\% & 0.40 & 0.8\% & 1.20 &1.47\% & 1.63& 3.55\% \\
P2 & 0.64 & 3.30\% & 0.20 & 5.45\% & 0.32 & 2.05\% & 0.43 & 0.92\% & 3.23 & 3.79\% & 2.51  & 5.46\%\\
\bottomrule  
\end{tabular}}
\caption{The average standard and ratio of standard and average.}
\label{tab:ratio_sum}
\end{table*}

\subsection{Anomaly analysis}
\label{ana}
In Table.~\ref{table:cls}, we can see the results in Evoprompt on TREC dataset is near 0. This is an anomaly, we can analyze the reasons behind Evoprompt's poor performance through specific examples on the TREC dataset. Our evaluation follows Evoprompt’s original prompt for the TREC dataset: \textit{Identity the inputs (explanations, entities, or humans) and use the outputs (numbers, descriptions, or entities) to answer the questions to make it easy understand for non-native English speakers.} The task classifies questions into ['Description', 'Entity', 'Expression', 'Human', 'Location', 'Number']. However, Evoprompt’s prompt introduces a misleading causal relationship between input and output and lacks sufficient labels. Evoprompt uses Alpaca-7B for prompt generation and evaluation in classification tasks, excluding GPT-3.5-turbo. Our tests on other white-box models like Llama2-7B showed stable performance Table.~\ref{table:llama_cls}. However, we analyze small-scale white-box models struggle with true semantic understanding, relying more on keyword extraction for classification. In contrast, GPT-3.5-turbo and GPT-4o-mini are more sensitive to prompt nuances, making them prone to misinterpretation if prompts are poorly optimized. Evoprompt’s prompts may mislead these models, causing errors, highlighting the need for transferable prompts—a strength of our approach.

\subsection{Adding Suffix}
\label{add_suffix}
In addition to demonstrating the effectiveness of our method, we also append suffixes directly after the initial prompts. These suffixes explicitly describe the type of perturbation present in the test data. We evaluate this approach across datasets from the three tasks on GPT-4o-mini, and the results are shown in Table.~\ref{tab:sum_suf}, Table.~\ref{table:sim_suf}, Table.~\ref{table:cls_suf}.

We can see that while intuitively, adding a suffix directly after the prompts seems like a reasonable approach, its effectiveness is limited. Our approach comprehensively outperforms adding a statement to initial prompts. Except for the C1 perturbation of the disturbed SST-5 dataset, where adding a statement slightly outperforms our method. We believe that this is because only adding suffixed may not be able to avoid the impact of perturbation, and the combination of original input and perturbation to explore the semantic space can obtain better results.

\begin{table*}[!htbp]
\centering
\small
\setlength{\tabcolsep}{2.5mm}
\begin{tabular}{c|c|c|c|c|c|c}
\toprule
\textbf{Methods} & \textbf{\shortstack{S1 \\ perturbation}} & \textbf{\shortstack{W3 \\ perturbation}} & \textbf{\shortstack{W1 \\ perturbation}} & \textbf{\shortstack{S2 \\ perturbation}} & \textbf{\shortstack{S3 \\ perturbation}} & \textbf{\shortstack{W1 \\ perturbation}} \\ 
\midrule
APE & 45.36	&40.96	&43.83	&44.42	&43.78	&44.43  \\
InstructZero & 33.16 &32.95 &30.12	&25.24	&20.16	&27.14  \\
INSTINCT & 24.72 &22.56	&24.44	&25.36	&18.63	&21.25 \\
\textbf{PGO} & \textbf{47.47} & \textbf{45.57} & \textbf{46.51} & \textbf{48.03} & \textbf{48.22} & \textbf{50.14}  \\ 
\bottomrule  
\end{tabular}
\caption{The SARI($\uparrow$) score obtained by the prompts generated by the APE,InstructZero, INSTINCT for the task where the text simplification task.}
\label{table:sim_ape}
\end{table*}
\begin{table*}
\centering
\small
\setlength{\tabcolsep}{13pt}
\begin{tabular}{c|c|c|c|c|c|c|c||c}
\toprule
\textbf{Type} & \textbf{Methods} & \textbf{CR} & \textbf{SST-2} & \textbf{MR} & \textbf{SST-5} & \textbf{TREC} & \textbf{AG'news}  & \textbf{Avg.}\\  \midrule
\centering \multirow{5}{*}{P1} & APE & 67.5 & 11.0 & 61.5 & 7.0 & 0.0 & 50.0 & 45.0 \\
 & InstructZero & 81.0 & 79.5 & \textbf{78.5} & 20.0 & 1.0 & 15.5 & 45.9 \\
 & INSTINCT & 58.5 & 35.0 & 33.5 & 2.0 & 0.0 & 31.5  & 26.8\\
 & \textbf{PGO} & \textbf{86.8} & \textbf{75} & \textbf{84.8} & \textbf{44.5} & \textbf{65.3} & \textbf{82.0} &\textbf{73.0} \\ 
\midrule
\multirow{5}{*}{P2} & APE & 73.4 & 13.2 & 62.0 & 12.6 & 0.6 & 48.0 & 35.0 \\
 & InstructZero & 74.2 & 82.8 & 76.2 & 18.6 & 0.6 & 15.2 & 44.6 \\
 & INSTINCT & 54.4 & 31.8 & 34.6 & 2.0 & 0.4 & 24.0  & 24.5\\
 & \textbf{PGO} & \textbf{83.5} & \textbf{83.7} & \textbf{81.8} & \textbf{42.5} & \textbf{59.1} & \textbf{83.5} &\textbf{72.4} \\ 
\bottomrule  
\end{tabular}
\caption{Average score($\uparrow$) of the prompts from different method(APE, InstructZero, INSTINCT) on six classification datasets on GPT-4o-mini. The table in the upper half is a perturbation of type P1, and the table in the lower half is a perturbation of type P2.}
\vspace{-15pt}
\label{table:cls_ape}
\end{table*}

\begin{table}
\centering
\large
\resizebox{0.45\textwidth}{!}{
\begin{tabular}{c|c|c|c|c|c|c}
\toprule
\textbf{Methods} & \textbf{S1} & \textbf{W3} & \textbf{W1} & \textbf{S2} & \textbf{S3} & \textbf{W1} \\ 
\midrule
Add-suff & 42.56 & 32.60 & 38.15 & 42.38 & 45.25 & 48.11  \\
\textbf{PGO} & \textbf{47.47} & \textbf{45.47} & \textbf{46.51} & \textbf{48.03} & \textbf{48.22} & \textbf{50.14}  \\ 
\bottomrule  
\end{tabular}}
\caption{SARI($\uparrow$) values of the instruction generated by PGO under two types of perturbations, P1 and P2, compared with adding suffix.}
\label{table:sim_suf}
\end{table}

\begin{table}
\footnotesize
\resizebox{0.45\textwidth}{!}{
\begin{tabular}{c|c|c|c|c}
\toprule
\textbf{Type} & \textbf{Method} & \textbf{Rouge-1} & \textbf{Rouge-2} & \textbf{Rouge-L} \\ \midrule
\multirow{2}{*}{P1} & Add-suff & 16.63 & 3.91 & 13.84 \\
& \textbf{PGO} & \textbf{18.40} & \textbf{4.15} & \textbf{14.81} \\ \midrule
\multirow{2}{*}{P2} & Add-suff & 16.45 & 3.76 & 13.73 \\ 
& \textbf{PGO} & \textbf{18.43} & \textbf{4.19} & \textbf{15.03} \\ 
\bottomrule
\end{tabular}}
\caption{The average of Rouge-1($\uparrow$), Rouge-2($\uparrow$) and Rouge-L($\uparrow$) of prompts generated by different methods on the text summarization task }
\label{tab:sum_suf}
\end{table}

\begin{table}
\footnotesize
\resizebox{0.45\textwidth}{!}{
\begin{tabular}{c|cc|cc|cc}
\toprule
\multirow{2}{*}{Method} &
  \multicolumn{2}{c|}{\textbf{CR}} &
  \multicolumn{2}{c|}{\textbf{SST-5}} &
  \multicolumn{2}{c}{\textbf{TREC}}  \\
  \cmidrule{2-7}
  & \shortstack{P1.Pert} & \shortstack{P2.Pert}  & \shortstack{P1.Pert} & \shortstack{P2.Pert}  & \shortstack{P1.Pert} & \shortstack{P2.Pert}  \\ \midrule
Add-suff  & 83.5 & 75.6 &  44.5 & 40.4 &  52.5 & 55.8  \\
\textbf{PGO}  & \textbf{86.8}  & \textbf{83.5} &  \textbf{44.5} & \textbf{42.5} &  \textbf{65.3} & \textbf{61.1}\\ \bottomrule
\end{tabular}}
\caption{The average of the scores($\uparrow$) of the instructions generated by PGO under two types of perturbations, P1 and P2, under different perturbations on the three datasets compared with adding suffix.}
\label{table:cls_suf}
\end{table}

\subsection{Cost Analysis}
\label{cost}
In this section, we evaluate the overhead of PGO in generating prompts. The primary overhead arises from the evaluation and generation processes during Optimization. The total cost is represented by the following relation: \( N \times (A + O) \), where \( N \) is the number of iterations, \( A \) represents the cost of the perturbation phase, and \( O \) represents the cost of the optimization phase. When calling the LLMs API, the cost is primarily determined by the number of tokens processed, including both input and output tokens. To estimate the cost of our method, we calculate the number of tokens required for executing tasks on three different datasets (XSum, Asset, and SST-5) across three types of tasks. This provides an understanding of the computational overhead associated with our approach. The results are in Table.~\ref{tab:cost}
\begin{table}
\footnotesize
\large
\resizebox{0.45\textwidth}{!}{
\begin{tabular}{c|c|c||c}
\toprule
\textbf{Datasets} &  \textbf{Perturb Phase} & \textbf{Optimize Phase} & \textbf{Total}  \\ \midrule
SST-5(P1)  & 0.0064M & 0.0194M & 0.0258M  \\
SST-5(P2)  & 0.0025M & 0.0369M & 0.0394M  \\
Asset(P1)  & 0.0289M & 0.1604M &  0.1893M \\
Asset(P2)  & 0.0554M & 0.8292M  & 0.8846M\\
XSum(P1) & 0.2793M & 0.5692M & 0.8485M   \\ 
XSum(P2) & 0.2006M & 0.8395M &  1.0401M \\ 
 \bottomrule
\end{tabular}}
\caption{The cost of PGO in three dataset under two kinds of perturbations}
\label{tab:cost}
\end{table}
From the results in the table, it can be observed that for relatively simple tasks (such as classification), the token consumption of the PGO algorithm is only 0.0258M. In contrast, for more complex tasks (such as text summarization), the token consumption remains at a low 0.0258M. After completing one round of PGO, the total token consumption is just 1.04M. This demonstrates that, once trained, PGO does not incur a large number of tokens. In comparison, Evoprompt consumes at least 4.20M tokens when converging on the SST-5 dataset, which has relatively simple input, significantly exceeding the token consumption of our method on XSum. This indicates that PGO offers substantial cost savings.

\begin{figure*}
    \centering
    \includegraphics[width=1\textwidth]{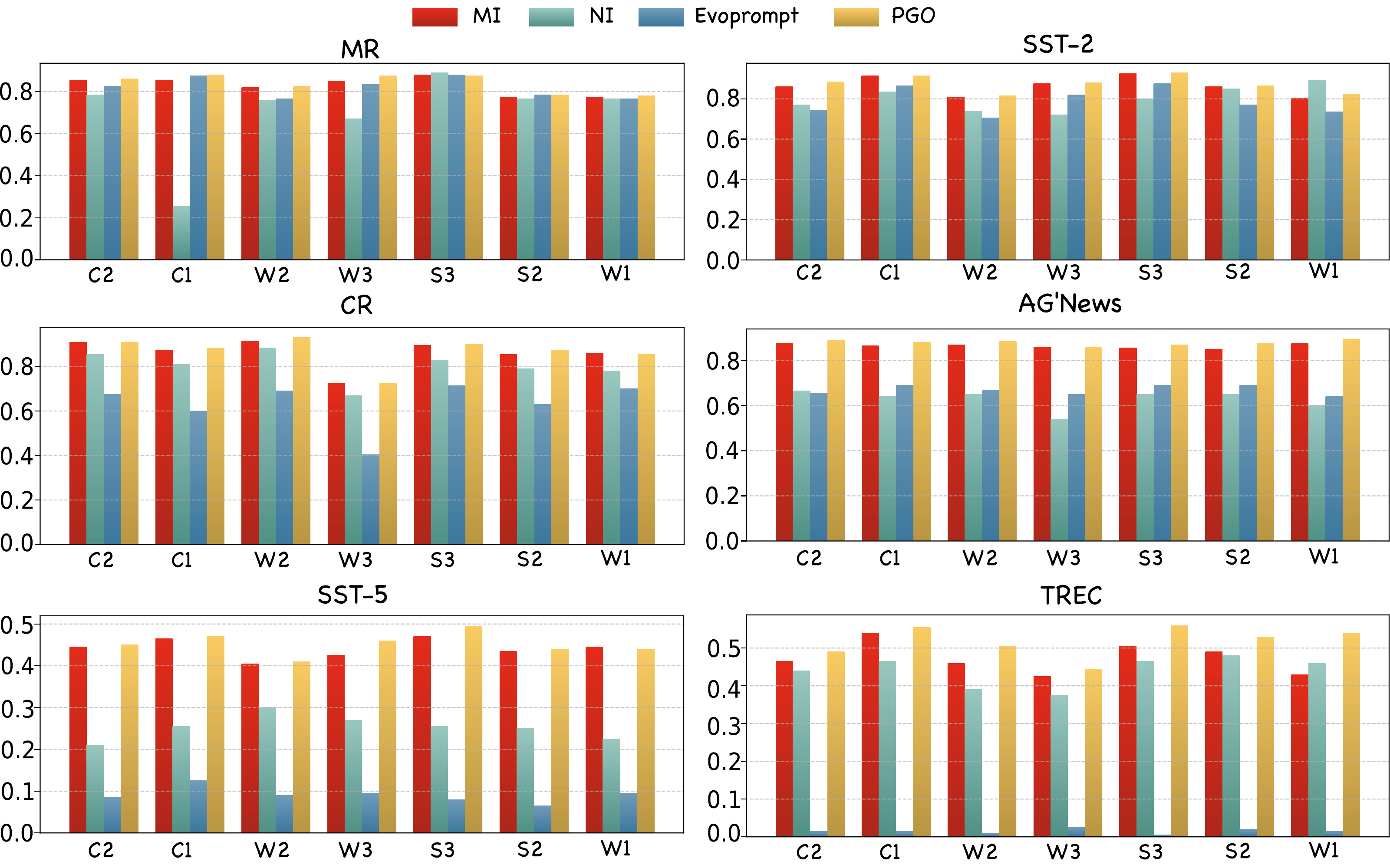}
    \caption{In classification, specific indicators of 6 datasets on different 7 kinds of perturbations}
    \label{fig:cls_de}
\end{figure*}

\subsection{Model Transferbility}
\label{trans}

\textbf{Effectiveness on GPT-4o-mini}.
To demonstrate the universality of our method in different LLMs on the baseline, we select the text summarization task and use \textbf{GPT-4o-mini} as the LLMs backbone. The prompts generated by PGO, along with those generated by several other methods, are evaluated under different perturbations using Rouge-1, Rouge-2, and Rouge-L metrics. The results obtained are averaged, shown in the Table.~\ref{tab:gpt4}. According to the experimental results, on different models, the prompts generated by PGO are still robust to different types of perturbations, and perform significantly better than other methods. In addition, we also tested the text simplification task and classification task on GPT-4o-mini. For the langugae-understanding task, We selected binary classification problem (CR) and multi-classification problem (SST-5) in sentiment classification datasets: as well as topic classification datasets. The specific experimental results are shown in the Table.~\ref{table:cls_gpt4} and Table.~\ref{table:sim_gpt4}. In summary, our method also performs well on GPT-4o-mini.

\begin{table}
\footnotesize
\resizebox{0.45\textwidth}{!}{
\begin{tabular}{c|c|c|c|c}
\toprule
\textbf{Type} & \textbf{Method} & \textbf{Rouge-1} & \textbf{Rouge-2} & \textbf{Rouge-L} \\ \midrule
\multirow{3}{*}{P1} & MI & 16.06 & 2.86 & 13.02 \\
& EvoPrompt & 16.88 & 2.62 & 14.44 \\
& \textbf{PGO} & \textbf{19.72} & \textbf{3.55} & \textbf{14.82} \\ \midrule
\multirow{3}{*}{P2} & MI & 16.14 & 2.82 & 12.98 \\ 
& EvoPrompt & 16.78 & 2.63 & 14.22 \\
& \textbf{PGO} & \textbf{19.26} & \textbf{3.71} & \textbf{15.12} \\ 
\bottomrule
\end{tabular}}
\caption{The average of Rouge-1($\uparrow$), Rouge-2($\uparrow$) and Rouge-L($\uparrow$) of prompts generated by different methods on the text summarization task on \textcolor[HTML]{229F69}{GPT-4o-mini}.}
\label{tab:gpt4}
\end{table}

 We plot the specific Rouge scores for each perturbation in the figure. The results show that, across all perturbations, our method demonstrates full robustness, outperforming the other prompt generation methods by a significant margin. Table.~\ref{table:sim_gpt4} and Table.~\ref{table:cls_gpt4} present the specific values of the prompt generated by PGO for the text simplification and classification tasks, respectively. In the text simplification task, PGO demonstrates optimal performance under various perturbations, with notable improvements in task scores, especially on GPT-4o-mini. Similarly, in the classification task, we selected several representative datasets, and the results show that our method retains robustness to perturbations across multiple datasets.

\begin{figure*}
    \centering
    \includegraphics[width=1\textwidth]{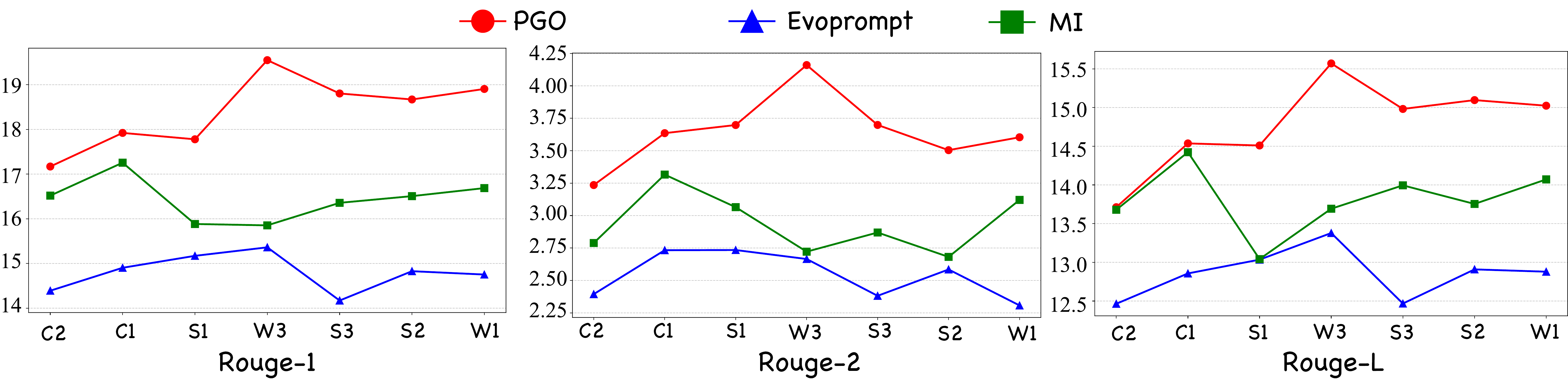}
    \caption{In the text summarization task, using \textcolor[HTML]{115EF8}{Llama2-7b} as backbone, the prompts produced by PGO and Rouge-1, Rouge-2, Rouge-L of the remaining methods on different disturbances.}
    \label{fig：llama_sum}
\end{figure*}

\textbf{Effectiveness on Llama2-7b}.
We also transferred the instructions generated by PGO to a white-box model to assess the effectiveness of our prompts. For the text summarization task, we used \textbf{Llama2-7b} to compute Rouge-1, Rouge-2, and Rouge-L scores. To evaluate the robustness of our optimized prompts under various perturbations, we present the results in the Table~\ref{tab:llama}. Our method still has optimal results. 
\begin{table}
\footnotesize
\resizebox{0.45\textwidth}{!}{
\begin{tabular}{c|c|c|c|c}
\toprule
\textbf{Type} & \textbf{Method} & \textbf{Rouge-1} & \textbf{Rouge-2} & \textbf{Rouge-L} \\ \midrule
\multirow{3}{*}{P1} & MI & 16.55 & 3.05 & 13.71 \\
& EvoPrompt & 14.82 & 2.62 & 12.78 \\
& \textbf{PGO} & \textbf{17.62} & \textbf{3.53} & \textbf{14.25} \\ \midrule
\multirow{3}{*}{P2} & MI & 16.35 & 2.85 & 13.88 \\ 
& EvoPrompt & 14.78 & 2.49 & 12.91 \\
& \textbf{PGO} & \textbf{18.98} & \textbf{3.74} & \textbf{15.17} \\ 
\bottomrule
\end{tabular}}
\caption{The average of Rouge-1($\uparrow$), Rouge-2($\uparrow$) and Rouge-L($\uparrow$) of prompts generated by different methods on the text summarization task on \textcolor[HTML]{115EF8}{Llama2-7b}.}
\label{tab:llama}
\end{table}

There are also additional details on the experiments conducted using Llama2-7b as the backbone for the text summarization task, as well as the experimental data for text simplification and classification. The results for the text summarization task are presented in the Figure.~\ref{fig：llama_sum}, where it is evident that PGO consistently outperforms baseline methods across all data items, as measured by Rouge-1, Rouge-2, and Rouge-L scores.

\begin{table}
\centering
\large
\resizebox{0.45\textwidth}{!}{
\begin{tabular}{c|c|c|c|c|c|c}
\toprule
\textbf{Methods} & \textbf{S1} & \textbf{W3} & \textbf{W1} & \textbf{S2} & \textbf{S3} & \textbf{W1} \\ 
\midrule
MI & 41.00 & 37.58 & 39.65 & 41.74 & 39.58 & 41.87  \\
EvoPrompt & 35.25 & 28.80 & 30.12 & 32.12 & 32.64 &  31.62  \\
\textbf{PGO} & \textbf{41.60} & \textbf{40.54} & \textbf{40.13} & \textbf{42.82} & \textbf{41.83} & \textbf{43.54}  \\ 
\bottomrule  
\end{tabular}}
\caption{SARI($\uparrow$) values of the instruction generated by PGO under two types of perturbations, P1 and P2, under different perturbations on Asset in \textcolor[HTML]{115EF8}{Llama2-7b}}
\label{table:sim_llama}
\end{table}

The SARI scores for the text simplification task on the Asset dataset are presented in the table. Similarly, the results demonstrate that the prompt generated by PGO outperform other methods across multiple types of perturbations.

Our experimental results on the classification task on the model of Llama2-7b are shown in the Table.~\ref{table:llama_cls}. We calculated the average scores of each dataset under each perturbation. It can be seen that on the white-box model, the overall indicators of all datasets are lower. However, under the guidance of the instruction generated by PGO, the results of Llama2-7b have optimal values under both P1 and P2 perturbations. Through the above experiments, it can be seen that our method has sufficient robustness on both white-box and black-box.

\begin{table}
\centering
\large
\resizebox{0.45\textwidth}{!}{
\begin{tabular}{c|c|c|c|c|c|c}
\toprule
\textbf{Methods} & \textbf{S1} & \textbf{W3} & \textbf{W1} & \textbf{S2} & \textbf{S3} & \textbf{W1} \\ 
\midrule
MI & 45.32 & 40.20 & 43.26 & 44.63 & 45.56 & 44.47  \\
EvoPrompt & 47.06 & 45.22 & 46.40 & 47.89 & 47.02 & 49.53  \\
\textbf{PGO} & \textbf{47.47} & \textbf{45.57} & \textbf{46.51} & \textbf{48.03} & \textbf{48.22} & \textbf{50.14}  \\ 
\bottomrule  
\end{tabular}}
\caption{SARI($\uparrow$) values of the instruction generated by PGO under two types of perturbations, P1 and P2, under different perturbations on Asset in \textcolor[HTML]{229F69}{GPT-4o-mini}}
\label{table:sim_gpt4}
\end{table}

\begin{table*}
\centering
\small
\setlength{\tabcolsep}{13pt}
\begin{tabular}{c|c|c|c|c|c|c|c||c}
\toprule
\textbf{Type} & \textbf{Methods} & \textbf{CR} & \textbf{SST-2} & \textbf{MR} & \textbf{SST-5} & \textbf{TREC} & \textbf{AG'news}  & \textbf{Avg.}\\  \midrule
\centering \multirow{4}{*}{P1} & MI & 31.0  & 36.8 &  31.8& 15.0 & 23.5& 48.5 & 31.1\\
 & NI & 34.0 & 40.3 & 37.0 & 24.5 &22.0 &74.0 & 38.6 \\
 & EvoPrompt &  37.5 & 39.3 & 43.0& 27.0 &17.3 &58.3 & 37.1 \\
 & \textbf{PGO} & \textbf{58.3} &\textbf{41.5}  &\textbf{44.5} &\textbf{27.3} &\textbf{24.5} &\textbf{76.8} &\textbf{45.5}  \\ 
\midrule
\multirow{4}{*}{P2} & MI & 28.4 & 33.0 & 31.2&17.5& 21.2 &55.8 & 31.8\\
 & NI & 31.9 & 36.1 &31.9 &24.1 &23.2 &76.8 & 37.3 \\
 & EvoPrompt & 32.8 & 30.9 & 38.8& 30.5 &18.3 &63.5  &35.8  \\
 & \textbf{PGO} &\textbf{34.1}  &\textbf{36.7}  &\textbf{39.0} &\textbf{36.0} &\textbf{23.9} &\textbf{78.2} &\textbf{41.3}  \\ 
\bottomrule  
\end{tabular}
\caption{Average score($\uparrow$) of the prompts from different method on six classification datasets using \textcolor[HTML]{115EF8}{Llama2-7b}. The table in the upper half is a perturbation of type P1, and the table in the lower half is a perturbation of type P2}
\label{table:llama_cls}
\end{table*}

\begin{table}
\footnotesize
\setlength{\tabcolsep}{0.2mm}
\resizebox{0.45\textwidth}{!}{
\begin{tabular}{c|cc|cc|cc}
\toprule
\multirow{2}{*}{Method} &
  \multicolumn{2}{c|}{\textbf{CR}} &
  \multicolumn{2}{c|}{\textbf{SST-5}} &
  \multicolumn{2}{c}{\textbf{TREC}}  \\
  \cmidrule{2-7}
  & \shortstack{w/o Pert.} & \shortstack{w/ Pert.}  & \shortstack{w/o Pert.} & \shortstack{w/ Pert.}  & \shortstack{w/o Pert.} & \shortstack{w/ Pert.}  \\ \midrule
MI  & 86.3 & 83.0 &  41.5 & 41.6 &  63.3 & 59.2  \\
EvoPrompt  & 84.5 & 82.6 &  2.25 & 3.7  &  2 & 1.6\\
\textbf{PGO}  & \textbf{86.8}  & \textbf{83.5} &  \textbf{44.5} & \textbf{42.5} &  \textbf{65.3} & \textbf{61.1}\\ \bottomrule
\end{tabular}}
\caption{The average of the scores($\uparrow$) of the instructions generated by PGO under two types of perturbations, P1 and P2, under different perturbations on the three datasets in \textcolor[HTML]{229F69}{GPT-4o-mini}.}
\label{table:cls_gpt4}
\end{table}

\section{Discussion}
\label{dis}
\subsection{Number of iterations}
In our optimization process, we set the number of iterations to five, as we observed that the prompts optimized by the LLMs achieved optimal performance at the fifth iteration. To illustrate this, we used the text summarization task, a computationally intensive text generation task, as an example. We tested the performance of prompts generated across six iterations (from 1 to 6), and the results are presented in the Experiments.

For each iteration, we evaluate the prompts across different perturbation types and calculate their average performance. The results show a consistent improvement in prompt performance during the initial rounds, with the optimal performance observed around the fourth or fifth iteration. This trend suggests that the LLMs effectively refines the prompts through iterative gradient-guided optimization and semantic space exploration, progressively approaching the optimal solution. However, beyond the fifth iteration, the performance of the prompts have a slightly declines. In order to speed up the inference experiment and reduce the cost, we choose the number of iterations to be 5
\subsection{Performance on the original dataset}

\begin{figure}
    \centering
    \includegraphics[width=0.45\textwidth]{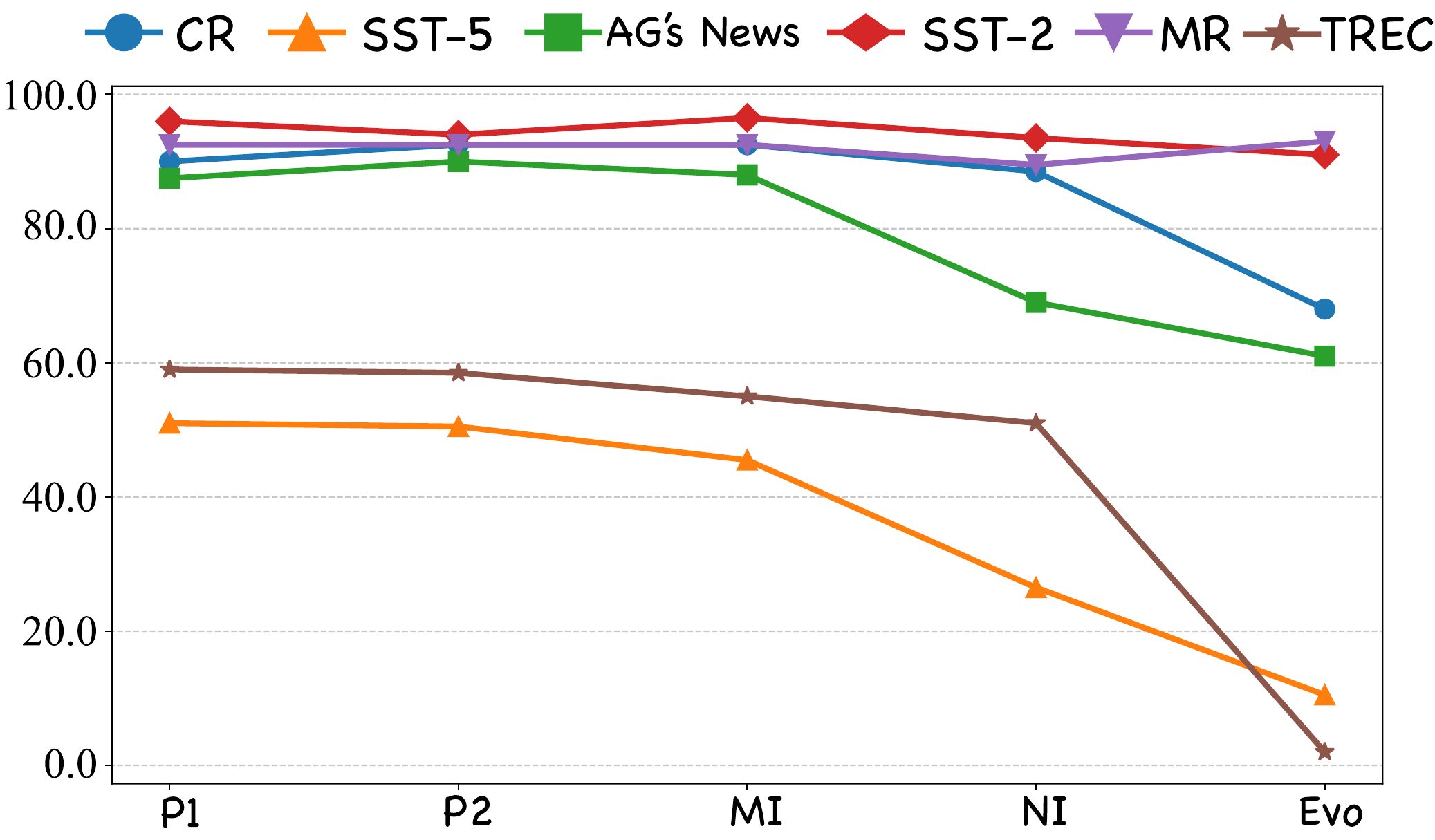}
    \caption{The performance of the prompts generated by PGO on undisturbed datasets of classification task.}
    \label{fig:cls_raw}
\end{figure}

In this section, we evaluate the performance metrics of prompts generated by PGO when applied to tasks on unperturbed datasets. This assessment demonstrates that our method is not only robust to perturbed data but also has greate performance on unperturbed data. The results for the classification tasks are illustrated in the Figure.~\ref{fig:cls_raw}. P1 and P2 represent different prompts generated by PGO for two types of perturbations. As shown in the Figure.~\ref{fig:cls_raw}, PGO achieves the best performance on several datasets (e.g., SST-5 and TREC). For other tasks, even when it does not outperform all methods, its performance is comparable to the best results (e.g., SST-2 and CR).

\begin{figure}
    \centering
    \includegraphics[width=0.45\textwidth]{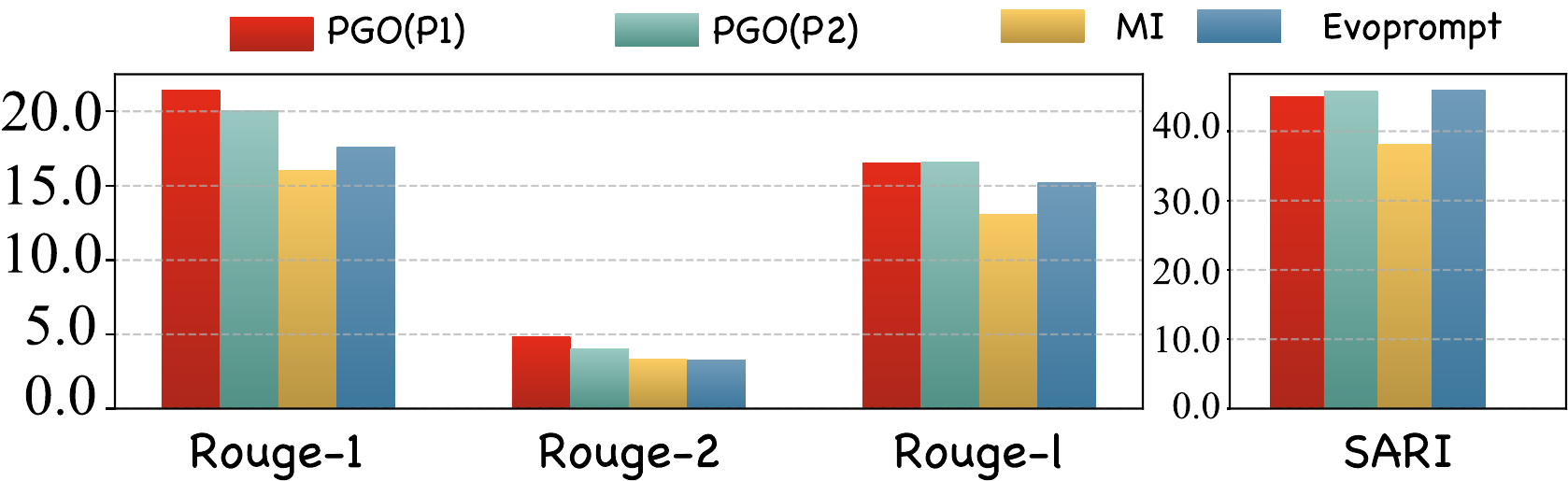}
    \caption{The performance of the prompts generated by PGO on undisturbed datasets of text simplification task and text summarization task.}
    \label{fig:ssraw}
\end{figure}

Similarly, the results for text generation tasks are presented in Figure.~\ref{fig:ssraw}, with the left figure illustrating the text summarization task and the right figure depicting the text simplification task. The experimental findings indicate that PGO also achieves the best performance on text generation tasks. Notably, in the text summarization task, its metrics are significantly higher than those of the second-best method. In conclusion, the prompts generated by PGO not only maintain robustness on perturbed datasets but also deliver strong performance on original, unperturbed datasets.

\section{Future Work}

\textbf{Exceptions Explore}: During our testing of the effects of different types of perturbations on task pairs, some intriguing observations emerged. While most perturbations negatively impacted task performance, certain perturbations surprisingly enhanced task execution. As shown in the Table.~\ref{tab:extra}, on the SST-5, AG's News, and Asset datasets, some perturbations appeared to unlock the latent potential of the large model, enabling it to perform better on the tasks.

We hypothesize that applying these perturbations to the input introduces diversity, encouraging the LLMs to engage its reasoning abilities rather than relying on specific representations. Additionally, such perturbations may influence the model's attention distribution, helping to resolve ambiguities in the input and ultimately enhancing task performance. This presents a highly valuable avenue for exploration. We could investigate a fixed perturbation strategy that consistently enhances the performance of LLMs when processing text inputs.

\begin{table}
\footnotesize
\setlength{\tabcolsep}{0.2mm}
\resizebox{0.45\textwidth}{!}{
\begin{tabular}{c|cc|cc|cc}
\toprule
\multirow{2}{*}{Pert.} &
  \multicolumn{2}{c|}{\textbf{SST-5}} &
  \multicolumn{2}{c|}{\textbf{AG'News}} &
  \multicolumn{2}{c}{\textbf{Asset}}  \\
  \cmidrule{2-7}
  & \shortstack{w/o Pert.} & \shortstack{w/ Pert.}  & \shortstack{w/o Pert.} & \shortstack{w/ Pert.}  & \shortstack{w/o Pert.} & \shortstack{w/ Pert.}  \\ \midrule
C1  & —— & —— &  —— & ——  &  38.12 & 47.64\\
C2  & —— & —— &  —— & ——  &  38.12 & 42.32\\
C3  & 36.0  & 37.5 &  78.3 & 79.3 &  38.12 & 39.47\\
S1  & 36.0 & 41.3  &  ——  &  —— &  —— & —— \\ 
 \bottomrule
\end{tabular}}
\caption{Examples of perturbations that have a positive effect on the dataset when performing a task. Where the Score of SST-5 and AG 'news is their prediction accuracy, and the Score of Asset is the SARI value.}
\label{tab:extra}
\end{table}

\section{Optimal Prompts}
\label{prompts}
\textbf{Iteration Prompts:} We optimized our prompt over five rounds, and the table below presents the final results for each iteration. We use the summarization task as examples. Since our approach aims to make LLMs better suited for the task while exploring the perturbed semantic space during optimization, the resulting prompts differ significantly from the original version. The results in the Table.~\ref{tab:Iterp}.

\textbf{Prompts for contrasting methods:} We publish the prompts that are optimal on different tasks after PGO generation and the prompts for Manual Instruction and Natural Instruction as baseline. Including classification(Table.~\ref{tab:cls_ins} and Table.~\ref{tab:cls_ins1}), text summarization(Table.~\ref{tab:sum_ins}), and text simplification(Table.~\ref{tab:sim_ins}).
From classification prompts, we can observe that for DA prompt, we observe that, due to the absence of explicit guidance, its prompt words and MI structure remain largely unchanged. This indicates that it fails to fully explore the semantic space to generate optimal prompts. For Evoprompt, it lacks targeted strategies to address perturbations. As a result, it can't help LLMs resist perturbations well. Our prompts not only explore a broader semantic space but also incorporate specific terms to mitigate the effects of perturbation. From language generation prompts, we can see that for Evoprompt contains excessive redundant information, such as "so readers can comprehend the important concepts and essential information.". This may affect the LLMs understanding. for simple iterative prompt, It suffers from the same problem as DA's prompts, which exhibits high similarity with MI, suggesting that the LLMs fails to explore a broader semantic space for generating more effective prompts. Our prompts actively navigate a richer semantic space, leading to more diverse and informative outputs.

\begin{table*}
\centering
\resizebox{\textwidth}{!}{
\begin{tabular}{l l p{15cm}}
\toprule
\textbf{Dataset} & \textbf{Method} & \textbf{Content}  \\
 \midrule 
\multirow{2}{*}{SST-2} & Manual Instruction & Please perform Sentiment Classification task. Given the sentence, assign a sentiment label from ['negative', 'positive']. Return label only without any other text. \\  
& Natural Instruction & In this task, you are given sentences from movie reviews. The task is to classify a sentence as "great" if the sentiment of the sentence is positive or as "terrible" if the sentiment of the sentence is negative. \\
&Evoprompt & Examine the movie reviews and classify them as either positive or negative. \\
&Data-Augmentation &Please assign a sentiment label ('negative' or 'positive') to the given sentence for the Sentiment Classification task. Submit only the label without any extra information. This task is expected to have better performance. \\
& \textbf{PGO(P1)} & $\cdot$ Decide if the text expresses a 'negative' or 'positive' sentiment without taking into account any extra details, even if there are mistakes in spelling and typos compared to the original text. 

$\cdot$ For this assignment, you will receive sentences that have been altered from movie reviews. Your job is to determine whether the sentiment of the sentence is positive or negative and classify it accordingly as either "positive" or "negative.\\
& \textbf{PGO(P2)} & $\cdot$ Conduct sentiment analysis by categorizing the sentiment of a sentence as \'negative\' or \'positive\'. Output only the sentiment label with no other information. 

$\cdot$ For this assignment, you will be provided with sentences taken from movie reviews. Your goal is to determine whether each sentence conveys a positive or negative sentiment by classifying it as either "positive" or "negative." The reviews may discuss specific characters, actions, critiques, relationships between characters, performances, or negative aspects of the works.\\

\midrule
\multirow{2}{*}{CR} & Manual Instruction & Please perform Sentiment Classification task. Given the sentence, assign a sentiment label from ['negative', 'positive']. Return label only without any other text. \\
& Natural Instruction & In this task, you are given sentences from movie reviews. The task is to classify a sentence as "great" if the sentiment of the sentence is positive or as "terrible" if the sentiment of the sentence is negative.  \\
&Evoprompt&  Analyze customer reviews and categorize each sentence as either ’positive’ or ’negative’.\\
&Data-Augmentation & Please complete the Sentiment Classification task by providing a sentiment label (either 'negative' or 'positive') for the given sentence. Only return the assigned label without any additional text. This task should have improved performance.\\
& \textbf{PGO(P1)} & $\cdot$ Analyze the sentiment of the modified text as either 'negative' or 'positive', disregarding any intentional spelling and grammar mistakes, and provide only the corresponding label as the result.

$\cdot$ For this assignment, you will receive original movie reviews. Your goal is to determine if a sentence has a positive sentiment by classifying it as "positive," or if it has a negative sentiment by classifying it as "negative.\\
& \textbf{PGO(P2)} & $\cdot$ Complete a Sentiment Classification task by analyzing a modified text with enhanced details and extra information. Provide a sentiment label of either \'negative\' or \'positive\', and only submit the label. 

$\cdot$ For this assignment, you will receive modified sentences from movie reviews. Your goal is to categorize a sentence as "positive" if it expresses positive sentiment or as "negative" if it expresses negative sentiment, regardless of any changes made to the text such as including negative adjectives or conveying different meanings using synonyms.
\\
\midrule
\multirow{2}{*}{MR}  & 
Manual Instruction & Please perform Sentiment Classification task. Given the sentence, assign a sentiment label from ['negative', 'positive']. Return label only without any other text. \\
& Natural Instruction & In this task, you are given sentences from movie reviews. The task is to classify a sentence as "great" if the sentiment of the sentence is positive or as "terrible" if the sentiment of the sentence is negative. \\
&Evoprompt &  Identify if a movie review is positive or negative by accurately categorizing each input-output pair into either ’positive’ or ’negative’.\\
&Data-Augmentation &Please provide a sentiment label ('negative' or 'positive') for the given sentence in the Sentiment Classification task. Submit only the label with no additional details. The task is anticipated to show improved performance.\\
& \textbf{PGO(P1)} & $\cdot$ Analyze the sentiment of the revised text given, ignoring any mistakes in spelling, typos, or edits made. Categorize the sentiment as either 'positive' or 'negative' without any additional details. 

$\cdot$ In this assignment, you will be given altered versions of movie reviews. Your goal is to classify each sentence as either positive for examining deeper themes and emotions, or negative for focusing solely on surface-level aspects of the films.\\
& \textbf{PGO(P2)} &$\cdot$ Complete the Sentiment Classification task by assigning a sentiment label of either 'negative' or 'positive' to the provided shortened sentence and return the label only. 

$\cdot$ For this task, you will be given distorted sentences from movie reviews. Your objective is to identify whether a sentence conveys a negative sentiment and label it as \'negative\', or if it conveys a positive sentiment and label it as \'positive\'.\\
\bottomrule
\end{tabular}}
\caption{Specific prompt of Manual Instruction(baseline), Natural Instruction, PGO in P1 and PGO in P2 in classification task.}

\label{tab:cls_ins}
\end{table*}

\begin{table*}
\centering
\resizebox{\textwidth}{!}{
\begin{tabular}{l l p{15cm}}
\toprule
\textbf{Dataset} & \textbf{Method} & \textbf{Content}  \\
 \midrule 
 \multirow{2}{*}{SST-5}  & 
Manual Instruction & Please perform Sentiment Classification task. Given the sentence, assign a sentiment label from ['terrible', 'bad', 'okay', 'good', 'great']. Return label only without any other text.\\
& Natural Instruction & In this task, you are given sentences from movie reviews. Based on the given review, classify it to one of the ﬁve classes: (1) terrible, (2) bad, (3) okay, (4) good, and (5) great.\\
&Evoprompt & Have your friend evaluate the movie they had just seen and provide a summary opinion (e.g. terrible, bad, okay, good, or great) to determine the sentiment of the movie review.\\
&Data-Augmentation & Perform the sentiment classification task by assigning a sentiment label of either 'terrible', 'bad', 'okay', 'good', or 'great' to the given sentence. Return only the label without any additional text.\\
& \textbf{PGO(P1)}  & $\cdot$ Perform a Sentiment Classification task by assigning a sentiment label from ['terrible','bad', 'okay', 'good', 'great'] to the modified sentence. Only include the label, no extra information needed.

$\cdot$ During this activity, you will be presented with sentences from movie reviews that have been modified to provide more general and subjective opinions on the movies. Your task remains the same: classify each review into one of the categories - terrible, bad, okay, good, or great - based on its sentiment and tone.\\
& \textbf{PGO(P2)} &$\cdot$  Perform Sentiment Classification. Given the altered text, assign a sentiment label from ['terrible', 'bad', 'okay', 'good', 'great']. Return the label only. 

$\cdot$ In this task, you are given sentences from movie reviews. Based on the given review, classify it to one of the five classes: (1) great, (2) good, (3) okay, (4) bad, and (5) terrible.\\
\midrule
\multirow{2}{*}{AG's News} 
& 
Manual Instruction & Please perform News Classification task. Given the news item, assign a label from ['World', 'Sports', 'Business', 'Tech']. Return label only without any other text.  \\
& Natural Instruction & In this task, you are given a news article. Your task is to classify the article to one out of the four topics "World", "Sports", "Business", "Tech" if the article"s main topic is relevant to the world, sports, business, and technology, correspondingly. If you are not sure about the topic, choose the closest option. \\ 
&Evoprompt & Please classify news articles into categories such as World, Sports, Business, or Tech. Output only the label without any additional text.\\
&Data-Augmentation & Please complete the Sentiment Classification task by providing a sentiment label (either 'negative' or 'positive') for the given sentence. Only return the assigned label without any additional text. This task should have improved performance.\\
& \textbf{PGO(P1)} & $\cdot$ Complete a News Classification assignment. Select a category from [\'World\', \'Sports\', \'Business\', \'Tech\'] for a news item that may have different spellings and minor grammar mistakes compared to the original text. Only provide the label without extra information.

$\cdot$ For this assignment, you will be provided with a news article. Your goal is to categorize the article into one of four topics: "World," "Sports," "Business," or "Tech" based on its main focus. If the article discusses trends in the U.S. stock market or gaming news, select either the "Business" or "Tech" category depending on the emphasis. If the article includes nonsensical or random words that do not make sense, choose the "Other" category.\\
& \textbf{PGO(P2)}& $\cdot$ Classify news by assigning a label from the options ['World', 'Sports', 'Business', 'Tech'] to each news item. Only provide the assigned label, without any extra text.

$\cdot$ For this task, you will receive a news article and your objective is to classify it under one of the four categories: "World", "Sports", "Business", "Tech" based on the article\'s main subject. If you are uncertain about the category, please select the most appropriate option. The Revised text includes extra details on different subjects and individuals like Tommy Tuberville and Major League Soccer, which are not referenced in the Original text. Additionally, the Revised text offers more precise information and examples, such as the mishandling of the conference\'s top position and the difficulties faced by Major League Soccer.\\
\midrule

\multirow{2}{*}{TREC} 
&  Manual Instruction &  Please perform Question Classification task. Given the question, assign a label from ['Description', 'Entity', 'Expression', 'Human', 'Location', 'Number']. Return label only without any other text. \\
& Natural Instruction & You are given a question. You need to detect which category better describes the question. Answer with "Description", "Entity", "Expression", "Human", "Location", and "Number".  \\
&Evoprompt & Recognize the inputs (explanations, entities, or humans) and provide the suitable outputs (numbers, descriptions, or entities) to answer the questions in a way that is understandable for non-native English speakers.\\
&Data-Augmentation & Please label the questions using one of the options: 'Description', 'Entity', 'Expression', 'Human', 'Location', 'Number'. Only provide the label, no extra text needed.\\
& \textbf{PGO(P1)} &$\cdot$ Determine the appropriate category for the text by providing only the label without any extra details. Choose from Description, Entity, Expression, Human, Location, or Number.

$\cdot$ Identify the appropriate category for the given text by selecting from "Description", "Entity", "Expression", "Human", "Location", and "Number" depending on the context.\\ 
& \textbf{PGO(P2)}&$\cdot$ Complete a Question Classification activity where you are provided with a shortened version of the original question without key words or phrases, and categorize it into one of the following labels: Description, Entity, Expression, Human, Location, or Number. Provide only the assigned label as the output. 

$\cdot$ Decide on the correct category for the text provided. Select from "Description", "Entity", "Expression", "Human", "Location", and "Number".\\
\bottomrule
\end{tabular}}
\caption{Specific prompt of Manual Instruction(baseline), Natural Instruction, PGO in P1 and PGO in P2 in classification task.}
\label{tab:cls_ins1}
\end{table*}

\begin{table*}
\centering

\resizebox{\textwidth}{!}{
\begin{tabular}{l  p{13cm}}
\toprule
\textbf{Method}  & \textbf{Content} \\
 \midrule 
Manual Instruction  & How would you rephrase that in a few words? \\
\midrule
Evoprompt & Reduce the core by reading or listening carefully to identify the main ideas and key points, so readers can comprehend the important concepts and essential information. \\\midrule
Data-Augmentation & How can you summarize that with better performance? \\\midrule
Simple-Iteration & How can the performance of the summary task be enhanced?\\
\midrule
\textbf{PGO(P1)} & Identify the main idea or central theme of the text. \\  \midrule
\textbf{PGO(P2)} &Determine the primary focus or central message of the text. \\\bottomrule

\end{tabular}}
\caption{Manual Instructions as the baseline and instructions with best performance generated by PGO (either P1 or P2) on Xsum.}

\label{tab:sum_ins}
\end{table*}

\begin{table*}
\centering
\resizebox{\textwidth}{!}{
\begin{tabular}{l  p{13cm}}
\toprule
\textbf{Method}  & \textbf{Content} \\
 \midrule 
Manual Instruction  & Simplify the text. \\
\midrule
Evoprompt & Rewrite the given sentence to make it more accessible and understandable for both native and non-native
English speakers. \\\midrule
Data Augmentation& Rewrite the text for improved performance.\\
\midrule
Simple-Iteration & Rewrite the text to improve its overall effectiveness.\\
\midrule
\textbf{PGO(P1)} & Rephrase the text using easier words without including any additional details. \\  \midrule
\textbf{PGO(P2)} &  Rephrase the text using more straightforward language and clearer wording to enhance understanding.\\\bottomrule
\end{tabular}}
\caption{Manual Instructions as the baseline and instructions with best performance generated by PGO (either P1 or P2) on Asset.}

\label{tab:sim_ins}
\end{table*}

\begin{table*}
\centering
\resizebox{\textwidth}{!}{
\begin{tabular}{l  p{13cm}}
\toprule
\textbf{Method}  & \textbf{Content} \\
 \midrule 
I1 & Can you provide a brief overview of the key points discussed in the text?\\
\midrule
I2 & Summarize the main ideas presented in the text in a clear and succinct manner. \\\midrule
I3 & Condense the key points and important facts from the text into a concise summary that captures the main ideas effectively\\
\midrule
I4 & Create a brief and compelling summary that encapsulates the main ideas and significance of the text\\
\midrule
I5 & Determine the primary focus or central message of the text. \\ \bottomrule
\end{tabular}}
\caption{Prompts in different Iterations in PGO.}

\label{tab:Iterp}
\end{table*}

\end{document}